\documentclass[journal]{IEEEtran}
\usepackage{cite}
\usepackage{amsmath,amssymb,amsfonts}
\usepackage{algorithmic}
\usepackage{algorithm}
\usepackage{graphicx}
\usepackage{textcomp}
\usepackage{url}
\usepackage{comment}
\usepackage{booktabs}
\usepackage{etoolbox}
\usepackage{multirow}
\usepackage{pifont}
\usepackage{makecell}

\usepackage[table,xcdraw,dvipsnames,svgnames,x11names]{xcolor}

\usepackage[hidelinks,colorlinks=true,linkcolor=Emerald,citecolor=cyan,urlcolor=black]{hyperref}

\usepackage{tikz,xcolor}

\definecolor{lime}{HTML}{A6CE39}
\DeclareRobustCommand{\orcidicon}{%
	\begin{tikzpicture}
	\draw[lime, fill=lime] (0,0) 
	circle [radius=0.14] 
	node[white] {{\fontfamily{qag}\selectfont \tiny ID}};
	\draw[white, fill=white] (-0.0625,0.095) 
	circle [radius=0.007];
	\end{tikzpicture}
	\hspace{-2mm}
}

\foreach \x in {A, ..., Z}{%
	\expandafter\xdef\csname orcid\x\endcsname{\noexpand\href{https://orcid.org/\csname orcidauthor\x\endcsname}{\noexpand\orcidicon}}
}

\begin{document}
\bstctlcite{IEEEexample:BSTcontrol}

\title{Hardware-Aware Learned Representation Compression for Distributed In-Sensor Vision}

\author{Chengwei Zhou, Abu Masum\orcidA{}, \IEEEmembership{Graduate Member,~IEEE}, Xuming Chen, Mehran~Moghadam\orcidB{}, \IEEEmembership{Graduate Member,~IEEE}, Sreetama Sarkar, Arnab Sanyal\orcidC{}, \IEEEmembership{Graduate Member,~IEEE}, Md Abdullah-Al Kaiser, M.~Hassan~Najafi\orcidD{}, \IEEEmembership{Senior Member,~IEEE}, Sercan~Aygun\orcidG{}, \IEEEmembership{Senior Member,~IEEE}, and Gourav Datta\orcidF{}, \IEEEmembership{Senior Member,~IEEE} 
        % <-this % stops a space
\thanks{Chengwei Zhou, Xuming Chen, Mehran Moghadam, M. Hassan Najafi, and Gourav Datta are with the Electrical, Computer, and Systems Engineering (ECSE) Department, Case Western Reserve University, Cleveland, OH 44106, USA (e-mail:
chengwei.zhou@case.edu; xuming.chen@case.edu; moghadam@case.edu; najafi@case.edu; gourav.datta@case.edu). \\ 
Abu Masum and Sercan Aygun are with the School of Computing and Informatics, University of Louisiana at Lafayette, Lafayette, LA 70503, USA (e-mail: c00591145@louisiana.edu; sercan.aygun@louisiana.edu). \\
Sreetama Sarkar is with the Department of Electrical and Computer Enginnering (ECE), University of Southern California, Los Angeles, CA 90089, USA (e-mail: sreetama@usc.edu). \\
Arnab Sanyal is with the Department of Electrical and Computer Engineering, The University of Texas at Austin, Austin, TX 78712, USA (e-mail: sanyal@utexas.edu). \\
Md Abdullah-Al Kaiser is currently with Apple; this work was completed while he was with the Department of Electrical and Computer Engineering at the University of Wisconsin–Madison, Madison, WI 53706, USA (e-mail: mkaiser8@wisc.edu). \\
An earlier version of this paper was presented at the 2026 IEEE Computer Society Annual Symposium on VLSI (ISVLSI)
2026 \cite{zhou2026oasis}.
}
%\vspace{-5pt}
%\vspace{-2.em}
}

%\author{Chengwei Zhou$^{1}$, Abu Kaisar Mohammad Masum$^{2}$, Xuming Chen$^{1}$, Mehran Moghadam$^{1}$, Sreetama Sarkar$^{3}$, Yuming Li$^{3}$, 
%Arnab Sanyal$^{4}$, Md Abdullah-Al Kaiser$^{5}$, M. Hassan Najafi$^{1}$, Sercan Aygun$^{2}$, Gourav Datta$^{1}$ \\
%$^{1}$Case Western Reserve University, Cleveland, USA \ $^{2}$University of Louisiana at Lafayette, Lafayette, USA \ $^{3}$University of Southern California, Los Angeles, USA \ $^{4}$The University of Texas at Austin, Austin, USA \ $^{5}$University of Wisconsin at Madison, Madison, USA  \\
        % <-this % stops a space
%\thanks{%This paper was produced by the IEEE Publication Technology Group. They are in Piscataway, NJ.
%}% <-this % stops a space
%\thanks{Manuscript received July  2026.%; revised August 16, 2021.
%}}

\markboth{IEEE Transactions on Emerging Topics in Computing,~Vol.~X, No.~X, XXXX~2026}%
{Shell \MakeLowercase{\textit{et al.}}: A Sample Article Using IEEEtran.cls for IEEE Journals}


\maketitle

\begin{abstract}
In-sensor computing reduces the cost of transmitting high-resolution image data by performing early-stage processing near the sensor. However, the logic chip integrated with a CMOS image sensor (CIS) is tightly constrained in compute and memory, limiting conventional deep neural network partitioning. We present OASIS, a distributed in-sensor vision framework that uses a lightweight encoder to generate compact, task-relevant representations before off-chip transmission. The encoder is trained end-to-end using task, entropy, and reconstruction objectives, while the decoder is used only during training. OASIS supports two complementary deployment paths. The first applies 4-bit quantization and Huffman coding while preserving the spatial structure required by classification and dense-prediction tasks. The second uses Sobol-based hyperdimensional computing (HDC) to transform the encoder latent into a fixed-dimensional binary hypervector for associative-memory classification. For the SwinViT-based VWW model, mapping a $3\times3\times8$ latent to a 64-dimensional hypervector provides an additional $1.77\times$ communication reduction with less than one percentage point of accuracy loss relative to the 128-dimensional configuration, yielding an overall $18{,}816\times$ reduction compared with raw 8-bit image transmission. We implement the digital near-sensor pipeline on an AMD Xilinx Zynq UltraScale+ FPGA and characterize it using direct board-level power measurements and Vivado post-implementation analysis, together with circuit-simulated CIS models and a 7-nm ASIC projection. Across visual wake-word classification, hand tracking, and eye tracking, OASIS reduces total system energy by approximately $2\times$--$4.5\times$ while maintaining competitive accuracy, demonstrating a practical hardware--algorithm co-design path for communication-efficient in-sensor vision.
\end{abstract}


\section{Introduction}

\IEEEPARstart{T}he rapid deployment of computer vision in applications including surveillance, autonomous systems, disaster response, and mobile devices has increased both the amount of visual data being captured and the computational demand placed on the corresponding inference pipelines. A conventional imaging system physically separates these two functions: a CMOS image sensor (CIS) first converts incident light into digital pixel values~\cite{fossum1997cmos}, after which the resulting image is transferred to a CPU, GPU, or dedicated accelerator for processing. For high-resolution and high-frame-rate operation, moving this pixel stream can impose substantial communication, throughput, and energy costs before any task-relevant information is extracted. 

A range of architectures has therefore attempted to move computation closer to image acquisition. Near-sensor systems place an accelerator adjacent to the sensor, either on the same platform or through heterogeneous integration~\cite{pinkhan2021jetcas,sony2020vision}; in-sensor systems move processing into the CIS periphery itself~\cite{chen2020pns, angizi2022pisa}; and in-pixel architectures perform selected computations directly within or immediately adjacent to the pixel array~\cite{datta2022p2m, scamp2020eccv, kaiser2024voltagecontrolledmagnetictunneljunction}. These approaches reduce the physical distance between sensing and computation, but several challenges remain. Most prior demonstrations operate on relatively simple datasets such as MNIST or low-resolution CIFAR-10, or implement only a small number of early neural-network operations. Supporting deeper multi-channel and multi-bit computations under the area, memory, and power constraints of the sensor remains substantially more difficult. Consequently, the activation tensor leaving the sensor can still be large, limiting communication savings and leaving a considerable fraction of the end-to-end computation on the external processor. 

We consider a heterogeneous architecture in which the CIS is coupled to a dedicated logic chip, following the general integration direction explored in prior sensor--compute systems ~\cite{gomez2022distributed,p2micrc}. Micro through-silicon vias ($\mu$TSVs) provide a high-bandwidth, low-energy connection between the sensor and this logic layer, allowing more processing to be performed before information reaches the comparatively expensive external interface. The central design problem is therefore not simply how many layers of an existing DNN can be moved onto the logic die. Instead, the sensor-side network should be explicitly designed to produce a representation whose spatial and channel dimensions are small enough to substantially reduce communication, while its computation and storage requirements remain compatible with the resource envelope of the integrated logic chip. This observation motivates OASIS, which treats sensor-side processing as a learned representation-compression problem. Rather than requiring the near-sensor encoder to independently solve the target application, we optimize it to generate a compact latent representation from which an off-chip model can recover the information necessary for the downstream task. This separation allows the sensor-side encoder to focus on aggressive dimensionality reduction while the more application-specific processing remains outside the sensor. The resulting architecture is particularly useful when the energy saved by reducing external communication and back-end processing exceeds the additional cost of computing the compact representation locally. 

The encoder is trained jointly with a reconstruction branch and a task-specific branch. Three complementary objectives shape the latent space: the downstream task objective encourages preservation of discriminative information; an entropy objective favors statistical distributions that can be represented efficiently after quantization; and an image-reconstruction objective discourages the encoder from discarding information that remains useful for describing the input. The reconstruction decoder is required only during optimization and is removed during inference, so its computational cost does not appear in the deployed sensor-side pipeline. In contrast to conventional image compression such as JPEG~\cite{wallace1991jpeg}, which is primarily designed around perceptual fidelity for human viewing, the learned OASIS representation is optimized around the requirements of machine inference. This also differs from approaches that directly execute inference on a conventionally compressed representation~\cite{torfason2018towards}: here, the compression front end itself is learned jointly with the target model. 

Our proposed architecture provides two ways to export this learned representation. The general path retains the spatial organization of the encoder output and combines low-bit quantization with Huffman coding, allowing the representation to support both classification and dense-prediction applications. For classification workloads, we additionally introduce an HDC path in which the encoder latent is converted into a single fixed-dimensional hypervector before it crosses the sensor-to-host interface. The latter removes the need to transmit a decodable spatial feature map and replaces the conventional task-specific classifier with associative-memory inference, providing a distinct operating point in the accuracy--communication--hardware trade space. Hardware evaluation is also essential because an arithmetic-operation count alone does not capture many costs that arise in an implemented sensor-side accelerator. Buffer organization, control logic, pipeline behavior, memory accesses, and interconnect activity can materially affect the energy consumed by the digital pipeline. We therefore implement the encoder and its post-processing datapath on an AMD Xilinx Zynq UltraScale+ platform and combine the resulting FPGA characterization with circuit-level simulation of the CIS front end. We further use the realized FPGA design as the basis for a $7$-nm ASIC energy projection representative of a future co-packaged logic die. This hardware-backed methodology provides a more concrete system-level evaluation than estimates based solely on per-MAC energy and operation counts. 

Across visual wake-word classification, hand tracking, and eye tracking, the structured quantization-and-Huffman path reduces the transmitted bit volume by as much as $11{,}985\times$ relative to raw $8$-bit image transmission. The HDC classification path provides an even more compressed operating point: for the SwinViT-based VWW configuration, mapping a $3\times3\times8$ latent tensor to a $64$-dimensional binary hypervector yields an overall $18{,}816\times$ reduction relative to the raw image. At the system level, OASIS reduces the modeled energy by approximately $2\times$--$4.5\times$ while maintaining competitive downstream-task performance.

\section{Background \& Related Works}

\subsection{In- \& Near-Sensor Computing}

Efforts to close the gap between image acquisition and computation fall 
along a spectrum defined by how deeply processing is pushed toward the 
photodiode. \emph{Near-sensor} designs place an accelerator adjacent to 
the sensor, either on the same board or 3D-stacked with the CIS, reducing 
but not eliminating the cost of moving raw pixels off the array. 
\emph{In-sensor} designs embed compute in the CIS periphery, while 
\emph{in-pixel} designs perform computation inside the pixel itself. The 
present work is best understood against the accuracy, efficiency, and 
generality limits that prior points on this spectrum have encountered, 
summarized in Table~\ref{tab:comparison}.

A recurring constraint is \emph{task and resolution scale}. Several 
influential in-pixel and in-sensor accelerators demonstrate their concepts 
on MNIST or low-resolution CIFAR-10 with shallow networks: SCAMP~\cite{scamp2020eccv} 
embeds compact CNNs on a pixel-processor array, MR-PIPA~\cite{mrpipa} 
integrates multilevel RRAM for in-pixel CNN execution, Senputing~\cite{xu2022} 
fuses sensing and computing for always-on inference with a two-layer MLP, 
and DPCE~\cite{dpce} and PiPSim~\cite{pipsim} target LeNet-class 
networks via direct-photocurrent computation and behavioral PIP modeling, 
respectively. While these works report strong per-operation efficiency, 
their evaluations do not extend to the high-resolution, multi-channel 
workloads representative of practical machine vision, and their shallow 
architectures limit the achievable activation compression.

A second constraint is the \emph{compute domain and reconfigurability}. Many high-efficiency designs rely on analog or mixed-signal in-pixel computation and emerging non-volatile devices, which constrains them to fixed early-stage 
operations (often a single convolution) and limits reprogrammability across 
tasks. P\textsuperscript{2}M~\cite{datta2022p2m} exemplifies the 
processing-in-pixel-in-memory paradigm at VWW-scale resolution 
($224^2$), but, like other analog approaches, performs only the first 
convolutional block in-pixel; the bulk of the network, and therefore the 
dominant share of system energy under Amdahl's law, still executes off-sensor, and the analog datapath offers limited flexibility across workloads.

A third, and frequently overlooked, constraint is \emph{evaluation 
methodology}. The efficiency figures reported by many in-/near-sensor systems derive from analytical per-MAC energy estimates that omit control logic, memory banking, pipeline stalls, and interconnect overhead, making 
cross-design comparison difficult and optimistic. Distributed sensor-logic 
frameworks such as~\cite{gomez2022distributed} improve realism with semi-analytical 
power models but still adopt conventional backbones that do not aggressively compress spatial and channel dimensions.

%These three limitations---restricted task/resolution scale, analog-bound and  weakly reconfigurable datapaths, and analytically-estimated rather than  hardware-measured energy---jointly motivate our approach: a lightweight,  fully \emph{digital} encoder that processes high-resolution inputs, aggressively compresses activations before the link, remains reprogrammable  across tasks, and is characterized with hardware-backed (FPGA-measured)  energy. Table~\ref{tab:comparison} positions our method against these prior  systems along resolution, technology node, network class, and measured  efficiency.

\subsection{Hyperdimensional Computing Engine}
Hyperdimensional Computing (HDC) has emerged as a compelling brain-inspired alternative computational paradigm, offering lightweight, energy-efficient data processing capabilities~\cite{ID-VSA_TVLSI2026}. Leveraging high-dimensional vector spaces, HDC has demonstrated exceptional efficacy across diverse domains, including healthcare~\cite{AMS-HD_2026}, IoT sensors~\cite{SenseHD_TCAD2026,SenHDC_TVLSI2025}, privacy and fault-tolerance applications~\cite{PP-HDC_DATE2024,HDC_Fault_Tolerance_TC2025}, cognitive modeling~\cite{kleykoSurveyII}, and few-shot learning~\cite{Seizure_HDC_TBME_2020}. Because it facilitates rapid, single-pass training and maps natively to low-power hardware, this framework is uniquely suited for real-time applications deployed on resource-constrained edge systems.
The beating heart of an HDC system encompasses the quasi-orthogonal high dimensional \textit{hypervectors (HVs)}, the atomic data structure in this paradigm, providing the capacity to represent scalar and complex symbolic information. 
Unlike conventional scalar architectures, HDC map disparate data modalities (e.g., spatial coordinates, alphanumeric characters, and temporal timestamps) into a unified and holistic, high-dimensional binary (logic-`0' and logic-`1') or bipolar ($\pm$1) vector format.
This distributed representation provides intrinsic fault tolerance; as the information is stochastically distributed across high dimensions, exhibiting robust resilience against hardware noise and bit-flip errors.

HVs are generated by comparing random source values with positional indicators or with scalar values over $D$ cycles. The encoding
process then applies a sequence of lightweight logical operations,
including \textit{permutation}, \textit{binding}, and \textit{bundling}.
Permutation reorders vector elements to preserve orthogonality,  binding combines multiple HVs through element-wise multiplication of bipolar values ($\pm$1)--or \texttt{XOR} in the binary computing case--and bundling merges multiple HVs into a single composite representation while maintaining data integrity.
Traditionally, HVs are generated using pseudo-random methods, such as using Linear-Feedback Shift Register (LFSR) in hardware, which may lead to limited orthogonality and degraded performance. To overcome this limitation, quasi-random sequences such as Sobol have been introduced~\cite{Aygun_Sobol_HDC_TCAD2025}, improving orthogonality and enhancing the reliability of neuro-symbolic encodings.

%Fig.~\ref{fig:hdc_model} illustrates the general workflow of an HDC model. Input features from training and test data are encoded into high-dimensional HVs through lightweight logic operations (e.g., \texttt{XOR}, \texttt{Add}, \texttt{Shift}, and \texttt{Permute}). During training, each encoded sample incrementally contributes to class representations, forming a distinct HV class for every category. This process yields the final deployable model. In inference, test samples are encoded in the same manner to generate query HVs. The query HV is then compared against all stored class HVs using similarity metrics, and the class with the highest similarity score is identified as the prediction.

\begin{figure*}[!t]
    \centering
\includegraphics[trim={120 193 165 240}, clip, width=1.\linewidth]{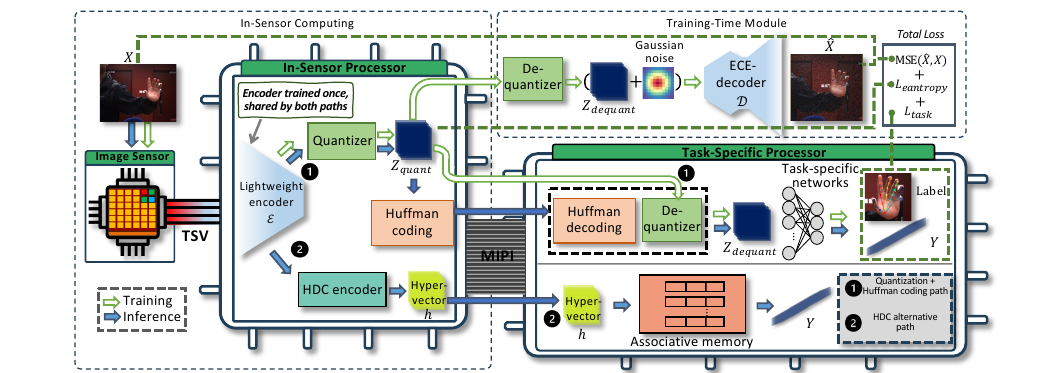}
    % \caption{
    % % Overview of the dual-branch autoencoder-based in-sensor computing architecture. The system integrates an \textit{image sensor} with an \textit{in-sensor processor} that implements a lightweight DNN backbone. Quantization and Huffman coding are applied at the processor's output to enhance data efficiency. An off-chip \textit{task-specific processor} handles specialized computations following the encoder. Additionally, a decoder branch is employed exclusively during training for reconstruction supervision, ensuring accurate learning and performance of the autoencoder.
    % %training \textbf{On-chip:} the backbone is deployed with an lightweight encoder to get compressed features. \textbf{Off-chip:} the branch is designed for task specific network. \textbf{QAT Module:} an additional branch is required for QAT, including a decoder for reconstruction supervision and to make a trade off for bit-distortion. 
    % %The processes with dash line arrows are only activated for QAT, and full line arrows are valid for both training and inference.
    % }
    \caption{Overview of the dual-branch autoencoder-based in-sensor computing architecture. The system integrates an \textit{image sensor} with an \textit{in-sensor processor} that implements a lightweight encoder $\mathcal{E}$. The output then is processed via two alternative paths: \ding{172} the default path, where quantization and Huffman coding compress the activations before transmission over MIPI to an off-chip \textit{task-specific processor} that performs Huffman decoding, de-quantization, and task-specific inference; and \ding{173} the HDC path, where the encoder latent is instead mapped on-chip to a fixed-dimensional hypervector $h$ and transmitted directly to an off-chip associative memory that predicts the label $Y$ via similarity search against stored class prototypes, eliminating the need for a separate classifier network. A decoder branch, active only during training, reconstructs the input from the de-quantized encoder output to provide reconstruction supervision.}
    \label{fig:framework}
    %\vspace{-3mm}
\end{figure*}

\section{Proposed Method}

\subsection{Dual-Branch Autoencoder-based Network} 
Figure~\ref{fig:framework} presents the overall OASIS architecture. The framework contains three principal learned components: a lightweight encoder $\mathcal{E}$ executed on the logic chip integrated with the CIS, a reconstruction decoder $\mathcal{D}$ that participates only during training, and an application-dependent processing branch that operates off-chip. This organization deliberately separates the representation-learning problem from final task inference. During deployment, the sensor-side encoder produces the representation that is communicated to the external processor, whereas the reconstruction decoder is completely removed from the inference datapath. 

\noindent\underline{\textit{On-chip Encoder Backbone}:} The objective of $\mathcal{E}$ is to transform the sensor input into a latent tensor whose size is substantially smaller than that of the original image while retaining information useful to the downstream application. We develop compact ResNet- and SwinViT-based encoders to study this design across convolutional and transformer architectures. Both variants are constructed so that spatial reduction occurs aggressively and channel growth is tightly controlled, limiting both intermediate memory demand and the volume of information that must ultimately leave the sensor. 

For the ResNet implementation, we adopt a compact residual structure inspired by~\cite{he2016deep}. The number of residual stages is reduced relative to a conventional ResNet configuration, and stride-$2$ operations progressively decrease the spatial resolution. The encoder therefore performs a large fraction of the required dimensionality reduction before its output reaches the sensor-to-host interface. For the transformer implementation, we construct a lightweight network based on the hierarchical Swin Transformer~\cite{liu2021swin}. Increasing the initial patch size from $4$ to $8$ reduces the number of tokens generated from a high-resolution input, while hierarchical downsampling further reduces the representation in subsequent stages. Channel expansion is deliberately limited in the early stages and reduced more aggressively near the encoder output. A final pooling layer is used where required to reach the target spatial dimension. These choices are motivated by the memory organization of the sensor-integrated logic die. A conventional backbone may gradually increase its feature dimension before achieving substantial spatial reduction, making it difficult to retain model parameters and intermediate activations in the available on-chip memories. In contrast, the compact OASIS encoders are designed so that both the model and working feature buffers can be accommodated locally, avoiding external DRAM accesses within the sensor-side processing stage. The resulting encoder output is subsequently handled by either the quantization-and-Huffman path or the HDC classification path described below. 

\noindent\underline{\textit{Decoder Branch}:} The decoder $\mathcal{D}$ provides reconstruction supervision during training but introduces no cost during deployed inference. The quantized encoder representation is first mapped back to its continuous-valued form and perturbed using simulated Gaussian noise before being supplied to the decoder. The noise injection exposes the decoder and encoder to perturbations associated with finite-precision representation and encourages the learned latent to remain stable under quantization. 

We organize $\mathcal{D}$ using an expand--contract--expand (ECE) structure. The first expansion increases representational capacity so that fine-grained information contained in the compressed latent can be recovered. The contraction stage then suppresses nonessential variation and encourages the network to preserve the features most useful for representing the input. Finally, the second expansion restores the spatial and channel dimensions needed for comparison with the original image. The reconstruction objective computed from this output provides an additional self-supervised signal during training, helping the encoder retain useful information even when its latent dimension is aggressively constrained. 

\noindent\underline{\textit{Off-chip Task-Specific Branch}:} For the structured compression path, the transmitted stream is first Huffman decoded and de-quantized by the task-specific processor. The resulting latent representation is then consumed directly by the downstream network. The exact back-end architecture depends on the application: VWW uses a classification head, whereas hand and eye tracking employ networks that preserve and process spatially organized information. Importantly, the reconstruction decoder is not involved in this path during inference. For the HDC alternative, the structured latent is instead converted on-chip to a hypervector and the conventional off-chip task-specific classifier is replaced by associative-memory similarity search, as described in Section~III-C.

\subsection{Training Methodology}

The complete network is optimized end-to-end so that compression does not become an isolated objective that destroys information needed by the application. We therefore combine reconstruction supervision, entropy regularization, and the downstream task objective as 
\begin{equation} \mathcal{L} = \mathrm{MSE}(X,\hat{X}) + \beta \mathcal{L}_{\mathrm{entropy}} + \gamma \mathcal{L}_{\mathrm{task}}, 
\end{equation} 
where $X$ is the input image, $\hat{X}$ is the output reconstructed by $\mathcal{D}$, and $\beta$ and $\gamma$ determine the relative contributions of the entropy and task losses, respectively. The reconstruction term constrains the compressed representation to retain information about the sensor input, while $\mathcal{L}_{\mathrm{task}}$ explicitly preserves features needed by the target application. 

The entropy term acts directly on the quantized encoder latent and is defined as \begin{equation} 
\mathcal{L}_{\mathrm{entropy}} = \max \left( \mathbb{E}_{Z_{\mathrm{quant}}\sim P} \left[ -\log_{2} P(Z_{\mathrm{quant}}) \right] - H_{\mathrm{ref}}, \,0 \right). \end{equation} 
Here, $P$ denotes the empirical probability distribution of the quantized representation, $P(Z_{\mathrm{quant}})$ is the probability assigned to a particular quantized encoder value, and $H_{\mathrm{ref}}$ specifies the target entropy. Following ~\cite{agustsson2017soft}, we estimate this discrete distribution using histogram statistics collected from the quantized latent. When its expected coding entropy exceeds $H_{\mathrm{ref}}$, the loss encourages the encoder to redistribute its outputs toward a smaller set of frequently occurring symbols. This makes the representation better suited to entropy coding without directly constraining every latent element to the same value. 

Quantization effects are incorporated during optimization using quantization-aware training (QAT)~\cite{jacob2018quantization}. Thus, the downstream task model and reconstruction branch observe the finite-precision representation that will be available during inference instead of being trained exclusively on full-precision activations. Because the rounding operation used by quantization is non-differentiable, we employ the straight-through estimator (STE)~\cite{bengio2013estimating}, approximating the backward derivative through the quantizer as 
\begin{equation} \frac{\partial \mathcal{L}}{\partial Z} \approx \frac{\partial \mathcal{L}} {\partial Z_{\mathrm{quant}}}, 
\end{equation} 
where $Z$ and $Z_{\mathrm{quant}}$ denote the full-precision and quantized encoder outputs. We use symmetric uniform quantization~\cite{li2023vit} and update the quantization range during training using a momentum-based procedure. For the branches that operate on a continuous representation, the quantized value is mapped back according to $Z_{\mathrm{dequant}} = Z_{\mathrm{quant}}\,q_{\mathrm{scale}}$, with 
$q_{\mathrm{scale}} = \frac{2\,\mathrm{max}}{2^{n}-1}$. Here, $\mathrm{max}$ represents the dynamically maintained magnitude of the encoder-output range and $n$ denotes the selected bit precision. This procedure exposes the encoder to the representation constraints encountered at deployment while retaining a differentiable optimization path through the remainder of the network.

\subsection{HDC as an Alternative Classification Path}\label{sec:hdc_details}

While quantization-aware training with Huffman coding  serves as our default compression mechanism, it compresses the encoder output 
\emph{element-wise}. The latent tensor retains its shape and each element is represented with a reduced effective bit-width. We additionally explore 
HDC as an \emph{alternative on-chip path} for classification tasks, in which the encoder latent is not quantized and transmitted as a feature map, but instead transformed into a single 
fixed-dimensional hypervector that crosses the sensor-to-host interface. 
In this configuration, the off-chip ``task-specific network'' is replaced 
by an associative-memory similarity search, and no decodable spatial feature map is required downstream. Consequently, this path is applicable only to classification workloads (e.g., VWW) and not to the dense-prediction tasks (hand and eye tracking) that require spatially structured outputs.

\noindent\underline{\textit{Latent-to-Hypervector Encoding}.} Let $\mathbf{z}\in\mathbb{R}^{m}$ 
denote the flattened encoder latent produced by $\mathcal{E}$ on the 
logic chip. Rather than quantizing $\mathbf{z}$, we map it into a 
$D$-dimensional hypervector space, where $D$ is chosen independently of 
the latent dimensionality $m$. This decoupling is the source of HDC's 
bandwidth behavior: the volume of data crossing the link is governed by 
$D$ rather than by the latent shape, so the achievable reduction is set 
directly by the chosen hypervector dimension (Table~\ref{tab:hdc_results}). Each 
normalized latent feature $\tilde{z}_{m}$ is quantized to a level index 
$k=\lfloor \tilde{z}_{m}(L-1)\rfloor$ and associated with a level 
hypervector $\vec{L}_{k}$ drawn from a precomputed set 
$\{\vec{L}_{\ell}\}_{\ell=0}^{L-1}$.

\noindent\underline{\textit{Sobol-Sequence Level Generation}.} The quality of an HDC model 
depends critically on the quasi-orthogonality of its level hypervectors. 
Pseudo-random generators such as LFSRs yield correlated vectors that 
degrade separability in high dimensions. Following~\cite{Aygun_Sobol_HDC_TCAD2025}, 
we instead generate level hypervectors from a low-discrepancy Sobol 
sequence, which improves orthogonality and yields more reliable encodings 
at a given $D$, directly benefiting the accuracy--dimension trade-off 
reported in Section~\ref{sec:hdc_result}.

\noindent\underline{\textit{Binding, Bundling, and Class Prototypes}.} The per-feature level 
hypervectors are combined through lightweight, logic-only operations: 
binding (element-wise multiplication for bipolar encodings, or XOR in the 
binary case) and bundling (accumulation followed by a sign operation) merge 
them into a single sample hypervector $\vec{h}=\mathrm{sign}\!\left(\sum_{m}\vec{v}_{m}\right)$. During training, 
each sample hypervector is accumulated into the prototype of its 
ground-truth class, and the final per-class prototypes are obtained by a 
sign operation over the accumulators (Algorithm~\ref{algo:hdc_train}). This 
single-pass training requires no gradient backpropagation and maps 
natively to low-power digital hardware.

\begin{figure}[!t]
%\vspace{-10pt}
\centering
\includegraphics[width=\columnwidth]{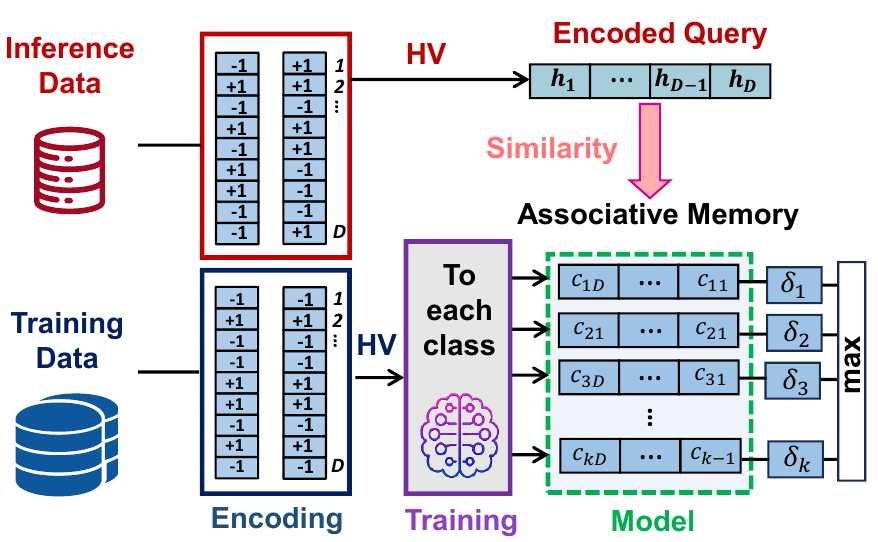}
%\vspace{-1em}
\caption{Overview of an HDC model: encoding, training, and classification via similarity search.}
\label{fig:hdc_model}
\vspace{-1em}
\end{figure}

\noindent\underline{\textit{Inference}.} At inference, a query latent is encoded into a 
hypervector via the same procedure and compared against all stored class 
prototypes using a similarity metric; the highest-scoring class is returned 
as the prediction. Because encoding and similarity search reduce to 
permute/bind/bundle/compare primitives over $D$-dimensional vectors, the 
entire classification path avoids the multi-bit MAC arithmetic of a 
conventional classifier head, and the stochastic, distributed nature of 
the representation provides intrinsic tolerance to bit-flip errors and 
hardware noise, an attractive property for in-sensor deployment. The 
overall encode--train--classify workflow is illustrated in Fig.~\ref{fig:hdc_model}.

\begin{algorithm}[t]
\footnotesize
\caption{Sobol-Based HDC Training}
\begin{algorithmic}[1]
\STATE \textbf{Input:} $\mathcal{D}=\{(\vec{x}_i,y_i)\}_{i=1}^{N}$, $D$, $L$
\STATE \textbf{Output:} $\{\vec{P}_c\}_{c=1}^{C}$

\STATE $\{\vec{L}_\ell\}_{\ell=0}^{L-1} \leftarrow \texttt{Sobol}(D,L)$
\STATE $\vec{A}_c \leftarrow \vec{0}\in\mathbb{R}^{D},\;\forall c$

\FOR{$i=1$ to $N$}
    \STATE $\tilde{\vec{x}}_i \leftarrow \texttt{Norm}(\vec{x}_i)$
    \FOR{$m=1$ to $|\tilde{\vec{x}}_i|$}
        \STATE $k \leftarrow \lfloor \tilde{x}_{i,m}(L-1)\rfloor$
        \STATE $\vec{v}_m \leftarrow \vec{L}_k$
    \ENDFOR
    \STATE $\vec{h}_i \leftarrow \texttt{sign}\!\left(\sum_m \vec{v}_m\right)$
    \STATE $\vec{A}_{y_i} \leftarrow \vec{A}_{y_i} + \vec{h}_i$
\ENDFOR

\FOR{$c=1$ to $C$}
    \STATE $\vec{P}_c \leftarrow \texttt{sign}(\vec{A}_c)$
\ENDFOR

\STATE \textbf{return} $\{\vec{P}_c\}$
\end{algorithmic}
\label{algo:hdc_train}
    \end{algorithm}

\subsection{Huffman Coding}
%Arnab Sanyal: Please write this.

\begin{figure}
    \centering
\includegraphics[width=0.45\textwidth]{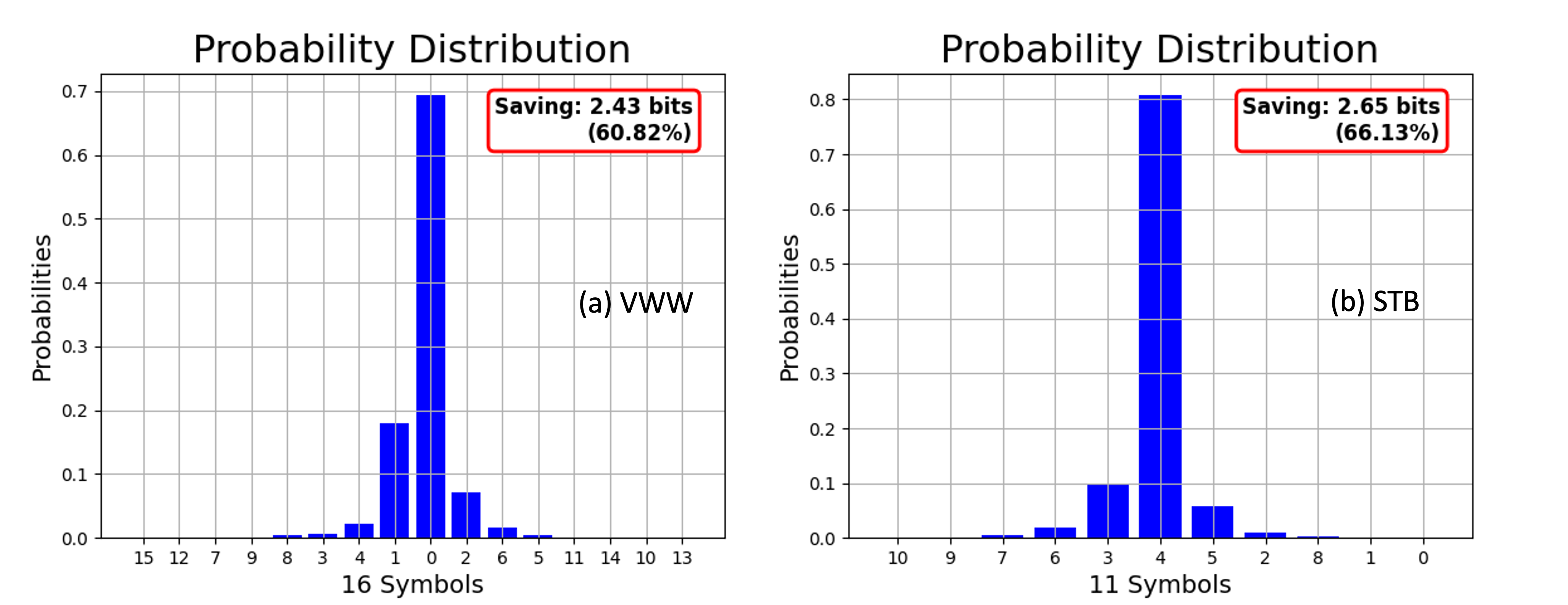}
    \caption{Probability distribution of the 4-bit quantized encoder output for input images from the evaluation sets of (a) the Visual Wake Words (VWW) dataset for image classification, and (b) the Stereo Hand Pose Tracking Benchmark (STB) dataset for hand tracking. 
    %\red{[Hassan: Fig.~1 has not been cross-referenced in the text. ]}
    }
    \label{fig:huffman}
    %\vspace{-3mm}
\end{figure}

After quantization, the encoder activations occupy a discrete alphabet whose symbols occur with markedly different probabilities. Figure \ref{fig:huffman} shows that the resulting distributions are concentrated around a relatively small subset of the available $4$-bit values rather than being uniformly distributed. We exploit this property using Huffman coding~\cite{huffenc}. A binary code tree is constructed from the measured symbol frequencies such that high-probability values receive short codewords and infrequent values are represented using longer sequences. Consequently, the average number of bits required for an encoder value can be substantially smaller than the nominal $4$-bit representation. Because this transformation is lossless, Huffman decoding exactly recovers the quantized symbol sequence at the receiver and therefore introduces no additional accuracy degradation beyond that already associated with quantization.

\noindent\textbf{Huffman Coding Is Not Applied to HDC Hypervectors.} 
Bit-wise Huffman coding does not benefit the HDC path. A binary source has only two symbols, $\{0,1\}$, so a Huffman tree always assigns each a 1-bit codeword regardless of skew, yielding no compression. This is consistent with our Sobol-based level hypervectors, which are designed to be low-discrepancy and quasi-orthogonal, pushing bit distributions toward uniformity rather than the skew Huffman coding requires. We therefore transmit the $D=64$ hypervector bits directly, without an entropy-coding stage.

%\vspace{-1mm}
\section{System Architecture}

The envisioned OASIS platform combines the imaging front end, sensor-side computation, and external task processing as shown in Fig. \ref{fig:framework}. The active-pixel sensor (APS) first converts the optical input into electrical signals and digitizes the resulting pixel values. These values are transferred through the local die-to-die interface to the integrated logic layer, where the lightweight encoder and the selected post-encoder processing path are executed. Only the compressed representation is subsequently communicated to the external task-specific processor.

\subsection{Active Pixel Sensor }

The APS employs column-parallel readout \cite{image_sensor}, with the pixels in each column sharing the analog-to-digital conversion circuitry located at the edge of the array~\cite{Kawahito2018}. Pixel voltages are therefore converted into digital values before entering the sensor-side compute layer. In the proposed heterogeneous organization, this digital stream is transferred directly to the logic die rather than being sent immediately through the conventional external camera interface. Locating the learned encoder next to the sensor also removes the need to execute a complete conventional image-signal-processing pipeline before task-oriented feature extraction. Moreover, because the downstream model is trained jointly with the sensor-side representation, the acquisition and processing pipeline can tolerate lower numerical precision than a general-purpose imaging path that must preserve high-fidelity pixels for arbitrary human viewing. Column-level low-power architectures, including successive approximation register (SAR) and multi-ramp ADCs, provide potential implementation options for trading conversion resolution against energy while maintaining the information required by the learned encoder.

\subsection{Efficient In-Sensor Processor}

The digital sensor-side accelerator uses a weight-stationary organization to reduce repeated movement of model parameters. Each processing element contains a quantized multiply--accumulate (MAC) datapath together with local storage used by the activation and weight operands. Because both OASIS encoders are intentionally compact, their parameters can reside entirely in on-chip BRAM, eliminating the need to fetch weights from an external DRAM during encoder inference. The two encoder families require different strategies for managing their intermediate data. In the SwinViT implementation, attention is restricted to local windows, and the feature map is processed one window at a time. Query, key, value, and attention-score storage therefore scales with the window size rather than with the full feature map. For the ResNet implementation, a small collection of scratch buffers is reused across successive residual blocks. This prevents memory capacity from growing in proportion to the number of network stages and allows intermediate activations to remain local to the accelerator. 

We exploit pipeline parallelism and loop-level concurrency throughout the implementation. For ResNet, the spatial convolution-kernel loops are fully unrolled, while accumulation across input channels is unrolled by a factor of four. Corresponding weight arrays are partitioned along the input-channel dimension so that the parallel MAC operations can access independent memory banks without creating port-contention bottlenecks. Batch-normalization, activation, and residual-addition stages are pipelined with an initiation interval (II) of $1$. A similar strategy is applied to the linear projections in the SwinViT encoder. The input-channel accumulation loops for the query, key, value, and output projections are unrolled by four, and the associated arrays are partitioned to provide the required memory bandwidth. Attention-score generation and the subsequent probability-weighted value accumulation are performed independently for each local window and are pipelined with $\mathrm{II}=1$. Intermediate buffers and inner accumulation dimensions are further partitioned to sustain this parallelism. 

For the structured output path, the $4$-bit quantizer and $16$-entry Huffman lookup table are also implemented as pipelined stages with $\mathrm{II}=1$. The lookup table is fully partitioned into registers so that the variable-length code corresponding to each input symbol can be obtained within a single cycle. A $64$-bit shift register accumulates these variable-length codes and emits packed $32$-bit output words. The HDC path is separately pipelined across latent features, with its hypervector accumulator fully partitioned so that all dimensions of the representation can be updated concurrently. 

On the AMD Xilinx Zynq UltraScale+ MPSoC implementation, the SwinViT encoder occupies approximately $37$K LUTs, $49$K flip-flops, $122$ DSP48E2 slices, and $474$ BRAM18K blocks, whereas the ResNet encoder requires approximately $21$K LUTs, $21$K flip-flops, $66$ DSP48E2 slices, and $177$ BRAM18K blocks. The post-encoder modules are much smaller: the Quantizer+Huffman implementation uses approximately $5$K LUTs and the HDC module approximately $3$K LUTs. Neither backend requires DSP resources, and each uses only two BRAM18K blocks. Thus, the hardware cost of selecting either post-encoder representation is small relative to the corresponding encoder accelerator.

\section{Hardware-validated System Energy Modeling}

A system-level evaluation of near-sensor processing must account for more than the arithmetic energy of the neural network. Sensor readout, data conversion, die-to-die communication, external transmission, digital encoder execution, and downstream task processing each contribute to the frame-level cost. We therefore combine \textit{Monte-Carlo circuit simulation} of the analog front-end, sweeping device mismatch and process, voltage, and temperature (PVT) variation across many randomized trials to capture worst-case as well as nominal behavior, with \textit{FPGA-based hardware measurements} of the digital compute pipeline, rather than relying exclusively on analytical per-operation estimates~\cite{gomez2022distributed,datta2022p2m,p2mglsvlsi}.

\vspace{-2mm}
\subsection{Energy Model Formulation}

The total energy per frame, $E_{Total}$, is expressed as:
\begin{align}
E_{Total} &= E_{Aps} + E_{Tsv} + E_{Inf} + E_{Enc} + E_{Back}
\end{align}
where $E_{Aps}$, $E_{Tsv}$, $E_{Inf}$, $E_{Enc}$, and $E_{Back}$ denote the energy contributions of the APS, TSV interface, sensor interface, encoder, and the task-specific back-end processor, respectively \cite{gomez2022distributed}. Note that the baseline non-in-sensor hardware does not incur $E_{Tsv}$, however, it incurs significantly higher $E_{Inf}$ that dominates $E_{Total}$.
%\vspace{-1mm}
\subsection{APS Energy Consumption}

The APS consumes per-frame energy in different operational modes, that is estimated as:
\begin{equation}
E_{Aps} = \left({E_{read/pix} + E_{ADC/pix}}\right)\cdot{N_{pix}}
\end{equation}
where $E_{read/pix}$, and $E_{ADC/pix}$ are the energy consumed in read-out and ADC conversion for each pixel respectively.
From our in-house circuit simulations in 22nm GlobalFoundries FDSOI technology, $E_{read/pix} + E_{ADC/pix} \approx 63.6$ \text{pJ} for 8-bit pixels.
%Another study \cite{} showed that a 10-bit SAR ADC in the same node has a Figure of Merit (FoM) of approximately $\approx 20$ \text{fJ/conversion-step}. The energy per conversion is given by $E_{\text{ADC}} = \text{FoM} \times 2^{\text{ENOB}}$. Assuming a nominal pixel bit-depth of $10$ in APS,
%$E_{\text{ADC}} = 20 \times 2^{10} = 20 \times 1024 = 20,480\,\text{fJ} = 20.48\,\text{pJ}$
%Hence, the total APS energy is
%\[
%E_{\text{Aps}} = 63.6 \times 50,\!176  = 3.19\,\mu\text{J}
%\]
%\vspace{-1mm}
\subsection{Communication Energy}
Data transfer between the different hardware components in our proposed system occurs through multiple communication interfaces, each with distinct energy characteristics.

\noindent{\underline{\textit{1) Through-Silicon Via (TSV) Energy}:}
TSV is used for high-bandwidth, low-energy connections between the sensor and on-sensor compute layer. The energy per transmitted byte for TSV communication for an input image is $E_{TSV}{=}A_{TSV}{\cdot}E_{Byte,TSV}$, where $A_{TSV}$ represents the transmitted data size in bytes, and $E_{Byte,TSV}{=}6.25$ pJ \cite{Vivet2020ISSCC} denotes the TSV energy per byte. 
%\begin{equation}
%E_{TSV}{=}A_{Size}{\cdot}E_{Byte,TSV}{=}224{\times}224{\times}3{\times}6.25 pJ{=}0.94 \mu J \notag
%\end{equation}
%where $A_{Size}$ represents the transmitted data size in bytes, and $E_{Byte,TSV}{=}6.25$ pJ denotes the TSV energy per byte.

\noindent{\underline{2) MIPI or other Wired/Wireless Interface Energy}:} The Mobile Industry Processor Interface (MIPI) is used to transfer processed data from the sensor to an external processor. The energy consumption is given by $E_{inf}=A_{Size} \cdot E_{Byte,inf}$, where $E_{Byte,inf}{=}100 pJ$ \cite{Choi2021MIPI}. Our autoencoder system significantly reduces $A_{inf}$, which dominates the total energy, as evaluated in Section~VI-B. While MIPI is standard for most sensor interfaces to off-chip hardware, remote processing scenarios often utilize wired (Ethernet, USB) or wireless (Wi-Fi, 5G) communication interfaces to transmit compressed data to cloud-based or edge processors. In such cases, $E_{Byte,inf}$ can be substantially higher, depending on the specific communication technology. Consequently, our distributed in-sensor computing approach, which minimizes data transmission requirements, becomes even more effective in reducing the overall energy consumption.
%\begin{equation}
%E_{inf} = A_{Size} \cdot E_{Byte,inf} = 256\times 100 pJ = 0.02 uJ
%\end{equation}
%where $A_{Size}=256$ is the dimension of our encoder output with size $4\times 4\times 16$ that ingests the input image of size $224\times 224\times 3$. This, coupled with our 4-bit quantized Huffman encoded encoder activation (which yields an effective bit width of 2.5), provides a compression factor of $\frac{224\times 224\times 3\times 8}{4\times 4\times 16\times 2.5}{\approx}1882\times$ with our proposed in-sensor computing framework. $E_{Byte,inf}=100 pJ$ is the interface energy per byte for the Mobile Industry Processor Interface (MIPI), which is standard for most sensor interface to off-chip hardware. Note that a non-in-sensor computing hardware would incur significantly higher $E_{Inf}=15 \mu J$, and would dominate the total energy. For remote processing, wired or wireless communication interfaces are used to transmit compressed data to cloud-based or edge processors \cite{Pinkham2021}. In this case, $E_{Byte,inf}$ would be even higher depending on whether the communication occurs over a wired (Ethernet, USB) or wireless (Wi-Fi, 5G) link, and further improve the efficacy of our distributed in-sensor computing approach.

\vspace{-2mm}
\subsection{FPGA-Based Encoder Energy Characterization} 
To obtain an accurate and empirically grounded estimate of encoder energy, we measure the power consumption of the complete inference pipeline, including the encoder backbone, quantizer, and Huffman encoder, directly on an FPGA prototype. This captures the full system behavior including control logic overhead, memory access patterns, pipeline stalls, and interconnect losses, which are difficult to account for in purely analytical models that estimate compute energy from per-MAC operation counts alone~\cite{datta2022p2m, angizi2022pisa}.

\noindent\underline{\textit{1) Power Measurement Methodology}}:
Power consumption is estimated using the post-implementation power analysis reports generated by Xilinx Vivado. Dynamic power is obtained from switching-activity-based estimates of the programmable logic (PL) design, while static power corresponds to the device leakage reported by the tool. The reported power values are derived from the final placed-and-routed design operating at the target clock frequency and are used to compare the relative energy efficiency of different encoder configurations. The frame-level encoder energy is computed as $
E_{Enc}^{FPGA} = (P_{active} - P_{idle}) \times T_{frame}$, where $P_{active}$ and $P_{idle}$ denote the measured active and idle PL power respectively, and $T_{frame}$ is the per-frame inference latency.
\noindent\underline{\textit{2) ASIC Energy Projection}}:
The FPGA measurement provides a verified upper bound on encoder energy. To project the energy for a custom 7nm ASIC implementation, which is representative of the envisioned co-packaged logic chip, we apply established FPGA-to-ASIC scaling factors. Based on comparative studies of identical designs implemented on both platforms~\cite{kuon2007measuring}, FPGA-to-ASIC energy ratios range from $14\times$ to $40\times$ depending on design complexity, with digital signal processing kernels (most analogous to our encoder) exhibiting ratios of approximately $20\times$ when scaling from 16nm FinFET FPGAs to 7nm ASIC technology: $E_{Enc}^{ASIC} \approx \frac{E_{Enc}^{FPGA}}{\alpha_{FPGA \to ASIC}}$, where $\alpha_{FPGA \to ASIC} \approx 20$ reflects the combined overhead of programmable routing and LUT fabric (${\sim}$7--10$\times$), technology node scaling from 16nm to 7nm (${\sim}$2--3$\times$), and generic BRAM versus custom memory macros (${\sim}$1.5--2$\times$)~\cite{kuon2007measuring}. Importantly, the FPGA prototype captures the \textit{complete encoder pipeline} including control logic, memory access patterns, and datapath overhead, effects that are absent in analytical per-MAC estimates. Even without ASIC scaling, the FPGA measurements provide a realistic and reproducible energy characterization that strengthens the credibility of our system-level claims.
We estimate the energy consumed by the task-specific back-end ($E_{Back}$) similarly using FPGA measurements.

\subsection{Huffman Coding/Decoding Energy} 
The energy associated with entropy coding is evaluated separately from the neural-network computation. Huffman encoding consists primarily of symbol-to-codeword lookup and bitstream packing, whereas decoding additionally requires recovery of the corresponding symbols from the variable-length representation. Using previously reported low-power entropy-coding implementations~\cite{bayar2018low} together with technology-scaling assumptions for the projected $7$-nm implementation, we use $E_{\mathrm{Huff\mbox{-}enc}} = 0.96~\mathrm{pJ/byte}$,  and $E_{\mathrm{Huff\mbox{-}dec}} = 1.15~\mathrm{pJ/byte}$. For an encoder payload of $A_{\mathrm{enc}}$ bytes, the corresponding coding contribution is therefore \begin{equation} E_{\mathrm{Huff}} = A_{\mathrm{enc}} \left( E_{\mathrm{Huff\mbox{-}enc}} + E_{\mathrm{Huff\mbox{-}dec}} \right). \end{equation} Although decoding has a slightly higher per-byte cost than encoding, the combined entropy-coding energy remains much smaller than the energy associated with transferring an equivalent amount of data through the external interface or repeatedly accessing off-chip memory.

%To assess the energy efficiency of the Huffman encoder and decoder, we analyze their core computational steps: frequency analysis, tree construction, symbol encoding, and decoding. Based on prior studies on low-power entropy coding implementations in 45nm CMOS technology \cite{bayar2018low}, and process scaling trends and energy efficiency improvements observed in modern 7nm ASIC designs, we extrapolate Huffman encoding energy to $E_{huff{-}enc}{=}0.96$ pJ per byte and decoding energy to $E_{huff{-}dec}{=}1.15$ pJ per byte in 7nm technology. Given that our system applies Huffman encoding to the quantized encoder output with data size $A_{enc}$, the total energy is estimated as $A_{enc}\cdot(E_{huff-enc}{+}E_{huff-dec})$. While decoding incurs a slightly higher energy cost than encoding due to additional tree traversal steps, the total Huffman energy significantly more efficient than direct DRAM access and MIPI data transfer, both of which consume ${\sim}100$ pJ energy per byte. %For the hand tracking task for AR/VR, $E_{Back}=xx \mu J$.
%Thus, the estimated energy consumption per forward pass is $1.74 \mu{J}$, considering only MAC operations and weight reads \cite{sinangil2020cim, ieee2023mac, sram2017}.

%By integrating energy estimation models across different hardware components, this formulation enables system-level energy evaluation and optimization for applications benefitting from in-sensor computing.

\section{Experimental Results}\label{sec:results}

\subsection{Task Performance}
%\textcolor{red}{Need add: training settings}
We evaluate our dual-branch approach on resource-constrained tasks pertinent to smart home and AR applications, specifically focusing on Visual Wake Words (VWW)\cite{chowdhery2019visual} classification, hand tracking, and eye tracking. We quantize the encoder output to 4 bits and train the entire network end-to-end for 200 epochs using AdamW with a cosine learning rate scheduler, setting $H_{ref}=0.7$. For VWW, we set $\beta{=}2$, $\gamma{=}2$, $\text{lr}{=}0.002$ and weight decay of 1e-5, where \textit{lr} denotes the learning rate. For hand tracking, we set $\beta{=}2$, $\gamma{=}4$, $\text{lr}{=}0.1$ and weight decay of 1e-5. For eye-tracking, we set $\beta{=}2$, $\gamma{=}4$, and train our network
%start from a pre-trained EyeNet~\cite{feng2022edgaze} model and train our EyeNet(ch=4) model with encoder output comprising 4 channels and quantized to 4 bits. Since this is a segmentation task, we do not require reconstruction using MSE. Our models are trained for 50 epochs 
using an Adam optimizer with an initial $\text{lr}$ of 1e-3 following the same configurations given in \cite{feng2022edgaze}. %\textcolor{red}{Sreetama: Please add the Eye tracking training settings similarly.}

\noindent\underline{\textit{1) VWW Classification}:}
\begin{figure}
    \centering
\includegraphics[width=1\linewidth]{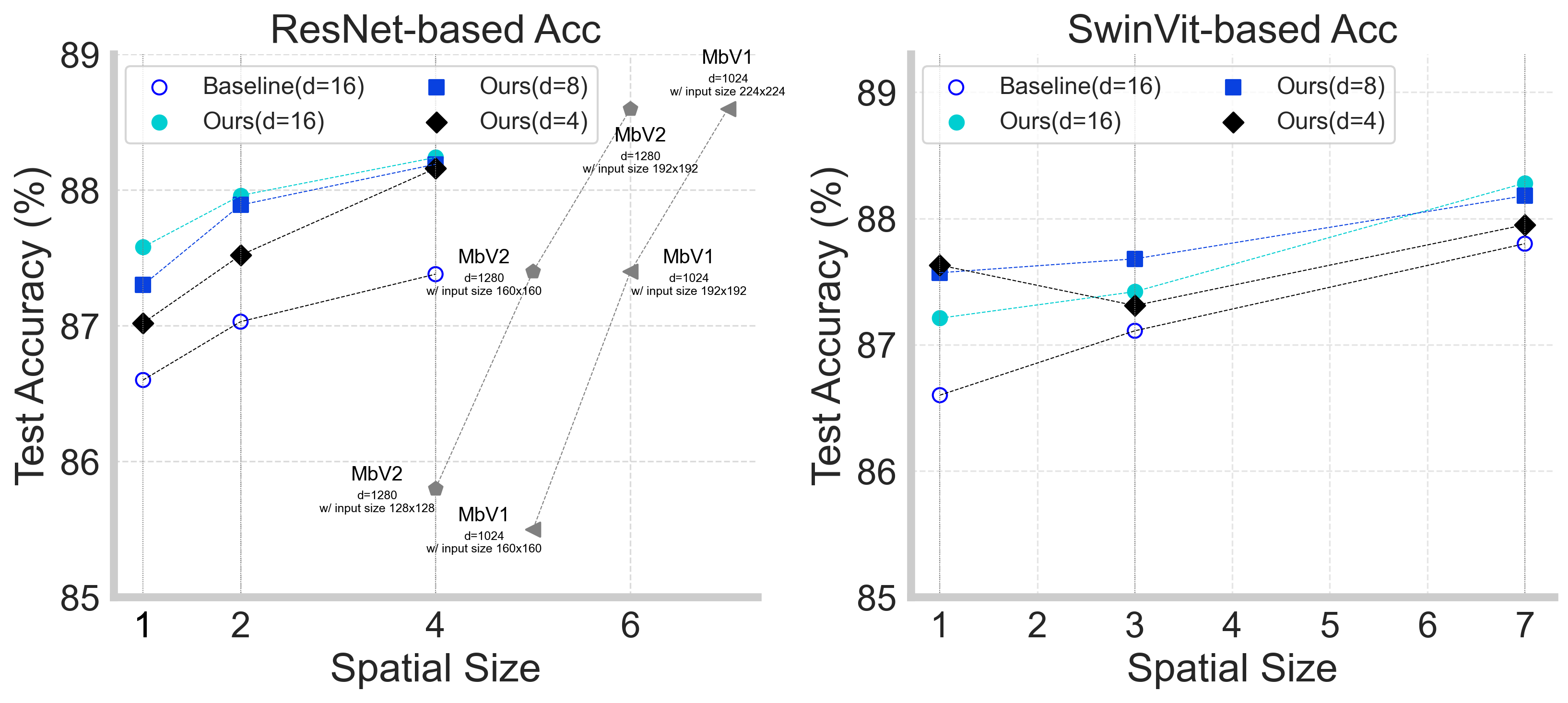}
    \vspace{-7mm}    \caption{Accuracy vs. Representation Dimension comparison on VWW classification task for (\textbf{Left}) ResNet and (\textbf{Right}) SwinViT based architecture, where $d$ denotes the channel dimension of the encoder output.}
    \label{fig:vww}
    %\vspace{-2mm}
\end{figure}
% To evaluate our approach under resource-constrained conditions, we conducted experiments on the image classification task using the Visual Wake Words (VWW) dataset. 
VWW evaluates whether an image contains a person and is commonly used as a benchmark for always-on and resource-constrained visual recognition~\cite{chowdhery2019visual}. We resize each input to $224\times224$ pixels and evaluate both ResNet- and SwinViT-based OASIS encoders so that the impact of the compression framework can be examined across convolutional and transformer backbones. \\
\noindent
%with dimension compression values $d$ and spatial compression sizes $s$.
\textit{a) ResNet-based Autoencoder}: Our compact ResNet-based encoder consists of three layers, each with two residual blocks. The initial convolutional module uses a \(7{\times}7\) kernel with a stride of 4 for aggressive downsampling. The channel dimensions of the three layers are configured as \([128,256,d]\), where \(d\) represents the encoder output channel dimension, which we vary to explore compression benefits. We apply uniform spatial downsampling with factors of 2 per layer, achieving a spatial dimension of \(s{\times}s\) at the encoder output. %through pooling. 
The ECE-decoder progressively reconstructs the feature maps, with layer dimensions \([d,256,128,256,128,64]\) and corresponding spatial scaling factors \([2{\textcolor{green}\uparrow}, 1, 2{\textcolor{green}\uparrow}, 2{\textcolor{red}\downarrow}, 2{\textcolor{green}\uparrow}, 2{\textcolor{green}\uparrow}]\), where \(2{\textcolor{green}\uparrow}\) and \(2{\textcolor{red}\downarrow}\) denote up-sampling and down-sampling by a factor of 2, respectively. As shown in Fig.~\ref{fig:vww}, even with substantial compression—e.g., \(d{=}16\) and \(s{=}1\) our method achieves significantly enhanced performance compared to the baseline configuration, which uses an encoder-only network under the same settings.
%of \([128, 256, 256]\) in the three layers), 
% with an accuracy increasement of ${>}8\%$. \\
\noindent
\textit{b) SwinViT-based Autoencoder}: Our SwinViT-based encoder divides the input image into patches of size \(8{\times}8\) and uses a window size of 7. The encoder comprises four stages (with two blocks per stage) and produces feature maps with dimensions \([64,128,160,d]\) and head number $[2,4,5,4]$ for the four stages. The output of ECE-decoder for each stage is with channel dimensions \([d,160,128,160,128,64]\), head number $[4,5,4,5,4,2]$ and spatial scaling factors \([2\textcolor{green}\uparrow,1,2\textcolor{green}\uparrow,2\textcolor{red}\downarrow,2\textcolor{green}\uparrow,2\textcolor{green}\uparrow]\). Fig.~\ref{fig:vww} shows that across various compression settings, our method consistently outperforms the baseline (an encoder-only network with channel dimensions \([64,128,160,16]\)), achieving significant performance gains while maintaining aggressive compression.
%in the transformer-based model.
% Our framework achieves a competitive classification accuracy of \textbf{XX\%} with only \textbf{XX} million parameters and \textbf{XX} GFLOPs computational cost, demonstrating superior efficiency compared to conventional CNNs (e.g., MobileNetV2: XX\% accuracy with XX MParams) and transformer-based approaches (e.g., NanoViT: XX accuracy with XX MParams). This validates our encoder’s capability to maintain task performance while drastically reducing computational overhead for in-sensor deployment.

\noindent\underline{\textit{2) Hand Tracking}:}
\begin{figure}
    \centering
    \includegraphics[width=1\linewidth]{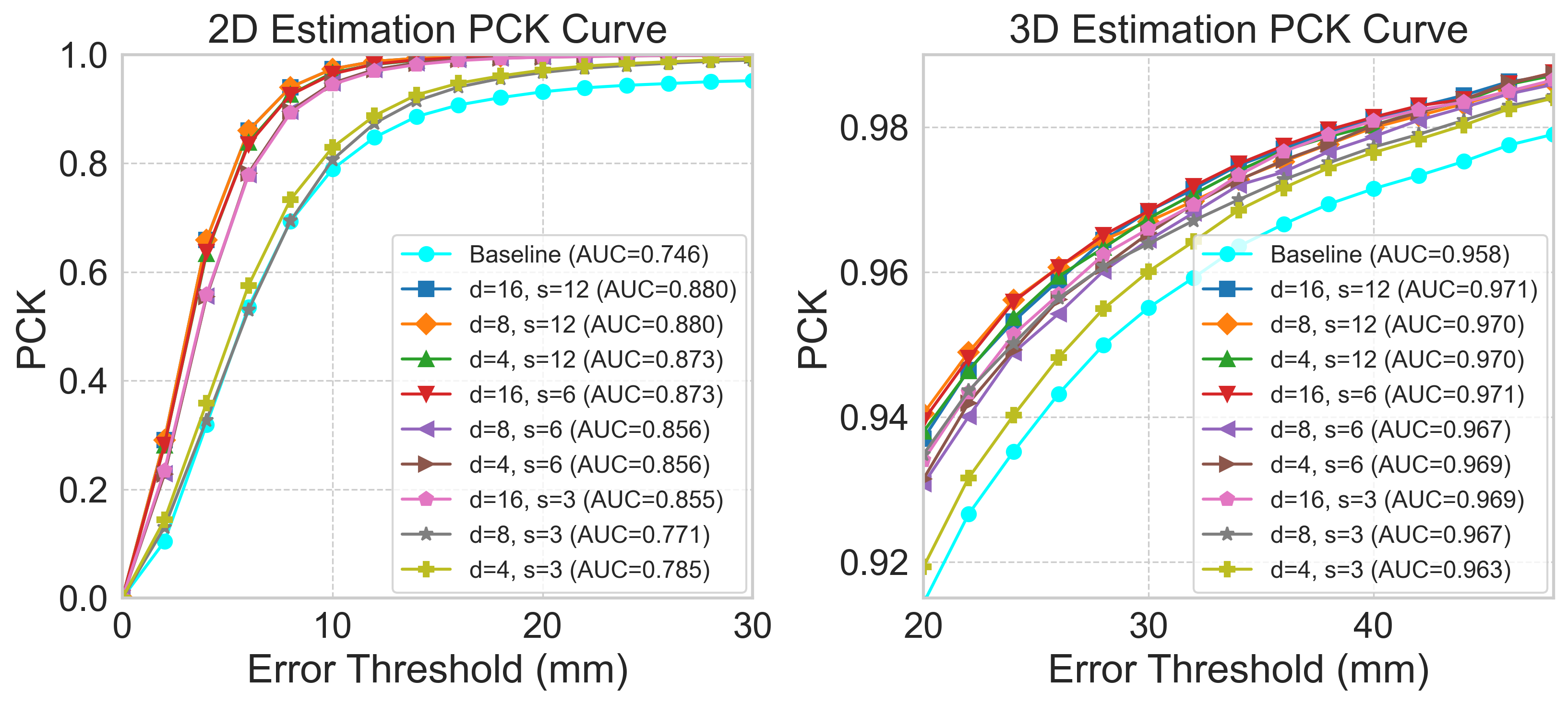}
    %\vspace{-7mm}
    \caption{\textbf{Left:} Comparison of 2D estimation AUC and PCK curves, truncated at 30mm, against the baseline.
    \textbf{Right:} Comparison of 3D estimation AUC and PCK curves, truncated within a deviation range of 20mm to 50mm. Baseline model is an encoder-only network with d$=$16, s$=$12.}
    \label{fig:ht}
    %\vspace{-4mm}
\end{figure}
%To evaluate the efficacy of our autoencoder system within an in-sensor computing scenario, we conduct a hand tracking task 
We use the Stereo Hand Pose Tracking dataset (STB) \cite{zhang2017hand}, 
%This dataset comprises 18,000 frames captured against six distinct backgrounds, featuring counting and random hand poses. 
which utilizes RGB data from the Intel RealSense F200 camera to estimate both 2D and 3D hand keypoints. The images were center-cropped and resized to \(96{\times}96\) pixels, removing redundant background while preserving the primary hand region.
%\noindent
%\textit{Network Architecture}: 
We adopt the KeyNet-F\cite{han2020megatrack} architecture, modifying the encoder to include four layers with channel dimensions \([32,32,32,64,d]\), where \(d\) represents the encoder output channel dimension, which we vary along with the spatial dimension $s{\times}s$ to explore compression benefits. 

%The spatial size of the encoder output is \(s{\times}s\).
We also develop an additional branch for the ECE-decoder with channel configurations \([d,32,64,32,32]\) and spatial scaling factors \([2{\textcolor{green}\uparrow},2{\textcolor{green}\uparrow},2{\textcolor{red}\downarrow},2{\textcolor{green}\uparrow},1]\). %where \(2{\uparrow}\) and \(2{\downarrow}\) denote upsampling and downsampling by a factor of 2, respectively. 
The task-specific network comprises a fused network, a keypoint heatmap network for generating 2D heatmaps, and a distance network, implemented to produce distance heatmaps for 3D keypoint estimation. 
%\noindent
%\textit{Performance Evaluation}: 
Figure~\ref{fig:ht} illustrates the impact of \(d\) and \(s\) on 2D and 3D hand keypoint estimation tasks. For 2D tasks, a higher channel count (\(d{=}16\)) and higher spatial resolution (\(s{=}12\)) yielded an Area Under the Curve (AUC) of 0.880, outperforming the baseline of 0.746. However, reducing the dimension count to \(d{=}4\) and \(s{=}3\) resulted in a lower AUC of 0.785. On 3D tasks, our networks achieved AUC values exceeding the baseline of 0.958. Even under aggressive compression ($d{=}16$ and $s{=}3$), the AUC remained high at 0.969.

\noindent\underline{\textit{3) Eye Tracking}:} Our eye tracking pipeline consists of two stages: (1) feature extraction for eye segmentation, followed by (2) gaze estimation. We use the EyeNet model as baseline \cite{feng2022edgaze}, a lightweight U-Net based segmentation model, and evaluate our approach on the OpenEDS \cite{palmero2020openeds2020} dataset in Table \ref{tab:eye_tracking}. To reduce data transmission bandwidth, we apply channel reduction from 32 to 4 at the encoder output of EyeNet using our proposed losses. Combined with 4-bit quantization, this achieves a 192$\times$ reduction in the encoder activation dimension as compared to 8-bit image inputs, and 24$\times$ reduction over encoder output size for baseline EyeNet with a mIoU degradation of only 1.4\%, as shown in Table \ref{tab:eye_tracking}. Further compression is possible through region-of-interest (ROI) prediction \cite{feng2022edgaze}, achieving up to $364\times$ overall reduction ($1.9\times$ reduction in input size via ROI). Since ROI prediction relies on previous frames, mIoU may improve over full sequences. However, due to limited annotated frames, evaluation is constrained, leading to mIoU degradation with ROI.

\begin{table}[t!]
    \centering
    \caption{Comparison of Eye segmentation results between EyeNet and our network on OpenEDS dataset w/ and w/o ROI prediction.}
    \label{tab:eye_tracking}
    \begin{tabular}{|l|c|c|c|}
    %\toprule
    \hline
        \textbf{Model} & \textbf{Inp. Dim.} & \textbf{Enc. Out Dim.} & \textbf{mIoU} \\ \hline
        %\midrule
        EyeNet & (1, 400, 640) & (32, 25, 40) & 0.988\\ \hline
        EyeNet (w/ ROI) & (1, 240, 560) & (32, 15, 35) & 0.967\\ \hline
        Ours & (1, 400, 640) & (4, 25, 40) & 0.974\\ \hline
        Ours (w/ ROI) & (1, 240, 560) & (4, 15, 35) & 0.939\\ \hline
    %\bottomrule
    \end{tabular}
    %\vspace{1mm}
    %\caption{Comparison of Eye segmentation results between EyeNet and our network on OpenEDS dataset w/ and w/o ROI prediction.}
    %\vspace{-7mm}
\end{table}

\begin{figure}
\centering
\includegraphics[width=0.98\linewidth]{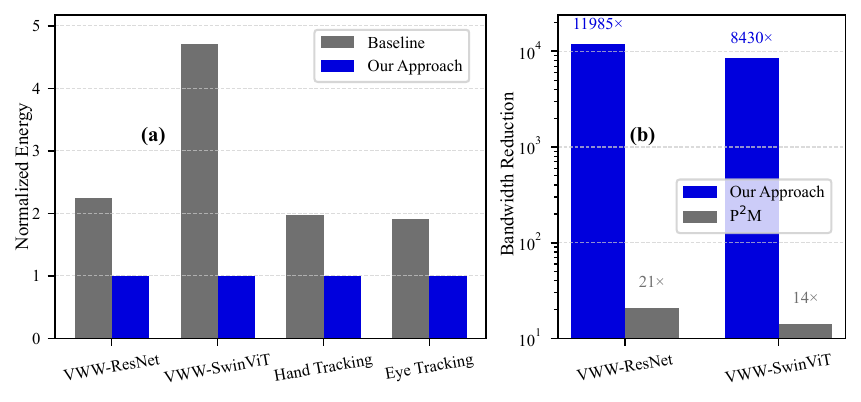}
    \caption{(a) Normalized energy consumption comparison between our approach and baseline across different vision tasks and models. (b) Comparison of bandwidth reduction with respect to the input image between our approach and P$^2$M. Our autoencoder-based system achieves substantial bandwidth reductions, which enables more efficient data transmission and contributes to the overall energy savings shown in (a). We report P²M bandwidth savings only for VWW tasks since these are the only common benchmarks in \cite{datta2022p2m}. For fair comparison, we use identical convolutional and pooling strides in the first layer for both ResNet and SwinViT models.}
\label{fig:energy_comparison}
%\vspace{-3mm}
\end{figure}

%\vspace{-1mm}
\subsection{Efficiency}

\noindent\underline{\textit{1) System Energy \& Communication}:}
%Bandwidth: Compared to a baseline non-in-sensor computing system, our proposed method reduces per-frame sensor bandwidth by $588\times$ for the VWW task at an input resolution of $224\times224$. For more complex tasks such as hand tracking, which require even higher input resolutions, the bandwidth reduction reaches 1240×. Furthermore, compared to an in-sensor computing system that implements only the first convolutional block in the sensor \cite{datta2022p2m}, our approach achieves a 28× bandwidth reduction at the same input resolution.
We compute the total energy consumed by our in-sensor computing hardware and baseline hardware using the energy model developed in Section~V. The energy components differ in several important aspects. The APS energy in Section~V-B remains identical for both hardware implementations.
%We compute the total energy consumed by our in-sensor computing hardware and baseline hardware using the energy model developed in Section IV. The energy components differ in several important aspects. The APS energy in Section IV-B remains identical for both hardware implementations. 
The in-sensor computing hardware incurs an additional TSV energy of $0.94\,\mu\text{J}$ for an input resolution of $224^2$ for internal data transmission between the sensor and logic chip, which the baseline hardware does not incur.

OASIS supports two complementary mechanisms for reducing off-chip communication. In the quantization-and-Huffman path, the encoder reduces the spatial and channel dimensions of the activation, after which 4-bit quantization and entropy coding further reduce the transmitted bit volume. For example, the configuration with a $4\times4\times4$ latent representation provides
\begin{equation}
R_{\mathrm{Quant+Huff}}
{=}\frac{224\times224\times3\times8}
     {4\times4\times4\times1.57}
\approx 11{,}985\times,
\end{equation}
where 1.57 is the measured effective number of bits per latent symbol
after Huffman coding. For classification workloads, the HDC path provides a more aggressive operating point. The SwinViT encoder produces a $3\times3\times8$ latent tensor with 1.57 measured effective number of bits, corresponding to
\begin{equation}
B_{\mathrm{latent}}{=}3\times3\times8\times1.57{=}113.57~\mathrm{bits}.
\end{equation}
Mapping the encoder representation alternatively to a 64-dimensional binary hypervector
reduces the sensor-to-host communication by
\begin{equation}
R_{\mathrm{HDC|latent}}{=}\frac{113.57}{64}{=}1.77\times.
\end{equation}
Relative to the raw $224\times224\times3$ 8-bit image, the resulting
communication reduction is
\begin{equation}
R_{\mathrm{HDC|raw}}{=}
\frac{224\times224\times3\times8}{64}{=}
18{,}816\times.
\end{equation}
%The HDC value is used as the headline communication result, whereas $11{,}985\times$ continues to denote the best quantization-and-Huffman operating point.
%Consequently, the MIPI interface energy is reduced by a factor of approximately $11,985/18,816$ compared to the baseline, where the MIPI energy is $99.5 \mu$J, which dominates the total energy consumption of $169.9 \mu$J. This significant reduction underscores the efficacy of our approach in minimizing energy consumption for edge-constrained applications. Similar trends are observed across other tasks and models. 
Accordingly, relative to raw-image transmission, Quant+Huff and HDC reduce the transmitted MIPI payload, and hence the modeled MIPI communication energy, by approximately $11{,}985\times$ and $18{,}816\times$, respectively. For this configuration, HDC therefore provides an additional $1.57\times$ communication reduction over Quant+Huff.”

%However, our in-sensor computing hardware achieves significantly lower MIPI interface energy due to the substantial activation dimension compression enabled by our autoencoder system. For instance, in the VWW task, our encoder network, with a spatial size of $4{\times}4$ and $4$ channels, achieves a dimension reduction factor of $\frac{224{\times}224{\times}3}{4{\times}4{\times}4}{=}2352$ with negligible accuracy loss. This reduction, combined with our 4-bit encoder output quantization (compared to the standard 8-bit unsigned representation for input images) and Huffman coding, which yields an effective bit-width of 1.57 bits from 4 bits, results in a total compression factor, or bandwidth reduction of $2352{\times}\frac{8}{4}{\times}\frac{4}{1.57}{\approx}11985$. Consequently, the MIPI interface energy is reduced by a factor of approximately $11985$ compared to the baseline, where the MIPI energy is $99.5 \mu$J, which dominates the total energy consumption of $169.9 \mu$J. This significant reduction underscores the efficacy of our approach in minimizing energy consumption for edge-constrained applications. Similar trends are observed across other tasks and models. 

As shown in Fig.~\ref{fig:energy_comparison}(a), our approach reduces the vision pipeline energy by ${\sim}2\times$ compared to baseline non-in-sensor systems across most tasks, with an exceptional ${\sim}4.5\times$ reduction for the SwinViT-based VWW task. This greater efficiency stems from our tiny SwinViT encoder, which significantly reduces the multi-head self-attention module size, drastically lowering the encoder energy by $13.4\times$ (our ResNet encoder reduces energy only by $1.15\times)$ and peak memory requirements. Unlike traditional DNN backbones that struggle to aggressively reduce activation dimensions and often require external DRAM access due to their larger memory footprint, our autoencoder-based approach optimizes the vision pipeline while respecting the strict compute and memory constraints of edge devices. Moreover, as shown in Table \ref{tab:comparison}, our approach achieves 22.7 TOPS/W, nearly twice the efficiency of prior works (11.49 TOPS/W in DPCE \cite{dpce}), while maintaining high accuracy. Unlike DPCE's binary multiplications, our method supports full multi-bit operations for complex convolution and self-attention functions. 
Compared to analog in-sensor computing systems such as P\textsuperscript{2}M~\cite{datta2022p2m}, our method extracts 
substantially more compression from the sensor-side logic. We note that 
this is not a controlled, like-for-like comparison: P\textsuperscript{2}M 
performs \emph{analog} computation of only the first convolutional block 
within the pixel array, whereas OASIS performs \emph{digital} multi-layer 
encoding on a logic chip after the TSV interface, in a different technology 
context. The two therefore occupy different points in the in-sensor design 
space rather than being direct substitutes. What the comparison does 
isolate, because both operate on the same $224{\times}224$ input resolution, is 
how much activation compression each \emph{paradigm} can achieve before the 
off-sensor link: P\textsuperscript{2}M reduces bandwidth by 
$\sim\!21\times$ via a single analog convolution, while our learned 
multi-objective encoder reduces it by $\sim\!11985\times$. The resulting 
ratio (Fig.~\ref{fig:energy_comparison}(b)) reflects the architectural advantage of a 
trained compression front-end over a single fixed analog layer, not a 
normalized per-layer or per-technology efficiency claim. 
%A fully normalized comparison would require re-implementing P\textsuperscript{2}M's analog front-end in our technology node and equalizing the number of in-sensor layers, which is outside the scope of this work.
%This makes our method particularly well-suited for edge-constrained applications.

\noindent\underline{\textit{2) FPGA Backend and Resource Efficiency}:}
To complement the system-level energy analysis, Table~\ref{tab:fpga_energy} reports the
hardware-measured latency and energy of the encoder accelerators and
the two post-encoder processing paths implemented on the AMD Xilinx
Zynq UltraScale+ MPSoC. The ResNet and SwinViT encoders consume
$33.8$~mJ and $45.4$~mJ per inference, respectively, whereas both
post-encoder backends operate in the sub-$2$~$\mu$J regime.
Consequently, the additional processing required after the learned
encoder contributes less than $0.01\%$ of the corresponding encoder
energy in all evaluated configurations. HDC is also more energy-efficient than Quant+Huff at the post-encoder
stage for both encoder families. For the ResNet output, the measured
backend energy decreases from $1.805$~$\mu$J with Quant+Huff to
$0.729$~$\mu$J with HDC, corresponding to a $59.6\%$ reduction. For
SwinViT, the backend energy decreases from $1.775$~$\mu$J to
$0.520$~$\mu$J, a $70.7\%$ reduction. Thus, although both backends
are negligible relative to the encoder itself, the HDC path provides
a consistently lower-energy classification backend.

Table~\ref{tab:fpga_power} further compares the FPGA resource utilization and measured
dynamic power of the individual hardware modules. The HDC backend
requires $3{,}258$ LUTs and $4{,}103$ FFs, compared with $5{,}031$
LUTs and $4{,}960$ FFs for Quant+Huff, corresponding to reductions of
$35.2\%$ and $17.3\%$, respectively. Both implementations require
only two BRAM18K blocks and no DSP resources. HDC exhibits a slightly
higher instantaneous dynamic power ($29$~mW versus $25$~mW for
Quant+Huff), but its substantially shorter execution time results in
lower total energy per inference. Overall, both post-encoder modules
introduce only a small hardware footprint relative to the ResNet and
SwinViT encoder accelerators.

\begin{table}[t]
\centering
\caption{Comparison of in-sensor computing approaches with varying DNNs, tasks, and CIS technology nodes.}
%\vspace{-2mm}
\label{tab:comparison}
\footnotesize
\setlength{\tabcolsep}{1.8pt}
\begin{tabular}{|l|c|c|c|c|c|c|}
\hline
\textbf{Method} & \textbf{Task} & \textbf{Res.} & \textbf{Tech.} & \textbf{Network} & \textbf{TOPS/W} & \textbf{Acc.} \\ \hline
Senputing \cite{xu2022} & MNIST & 28² & 180nm & 2-layer MLP & 4.7 & 93.76\% \\ \hline
SCAMP \cite{scamp2020eccv} & MNIST & 256² & 180nm & 2-layer CNN & 0.535 & 93.0\% \\ \hline
MR-PIPA \cite{mrpipa} & MNIST & 256² & 180nm & 3-layer CNN & 1.89 & 97.26\% \\ \hline
DPCE \cite{dpce} & CIFAR10 & 32² & - & LeNet-5 & 11.49 & 87.20\% \\ \hline
PIPSIM \cite{pipsim} & CIFAR10 & 64² & 45nm & LeNet-5 & 4.12 & 90.05\% \\ \hline
P$^2$M \cite{datta2022p2m} & VWW & 224² & 22nm & MobileNetV2 & 0.4 & 84.3\% \\ \hline
\textbf{Ours} & VWW & 224² & 22nm & {Tiny SwinViT} & \textbf{15.5} & {88.2\%} \\ \hline
\textbf{Ours} & Hand Track & 96² & 22nm & {KeyNet-F} & \textbf{22.7} & {96.9\%} \\ \hline
\multicolumn{7}{l}{$^*$\scriptsize{1 OP = 1 MAC between weight and input activation in a DNN.}}
\end{tabular}
%\vspace{-5mm}
\end{table}

\noindent\underline{\textit{3) Latency}:} The communication reduction provided by OASIS also directly affects
end-to-end latency. In conventional vision pipelines, APS readout and
subsequent transmission of the image or intermediate activations over
MIPI can represent a substantial portion of the processing delay.
OASIS instead transfers the raw sensor output over the high-bandwidth
TSV interface and communicates only the compact learned
representation through MIPI. The TSV interface provides approximately
$200\times$ greater bandwidth than MIPI~\cite{Vivet2020ISSCC,Choi2021MIPI}, making the local
sensor-to-logic transfer comparatively small.

The two OASIS paths provide different communication operating points.
Quant+Huff reduces the transmitted bit volume by up to
$11{,}985\times$ while retaining the spatially structured latent
required by both classification and dense-prediction tasks. For
classification, HDC further reduces the transmitted representation to
a fixed-dimensional binary hypervector, providing up to an
$18{,}816\times$ reduction relative to the raw image.

The FPGA measurements in Table~III show that the post-encoder
processing latency is negligible relative to encoder execution for
both paths. Quant+Huff requires $0.06944$~ms and $0.06827$~ms for the
ResNet and SwinViT encoder outputs, respectively, whereas HDC requires
only $0.03242$~ms and $0.03156$~ms. HDC therefore reduces the
post-encoder processing latency by approximately $53.3\%$ for ResNet
and $53.8\%$ for SwinViT relative to Quant+Huff. In all cases, these
latencies are orders of magnitude smaller than those of the encoder
accelerators, confirming that neither compression path introduces a
meaningful processing bottleneck.

\begin{table}[t]
\centering
\caption{Hardware-measured latency and energy of the proposed encoder, Quant+Huff, and HDC modules implemented on a Xilinx Zynq UltraScale+ MPSoC FPGA.}
\label{tab:fpga_energy}
\footnotesize
\setlength{\tabcolsep}{5pt}
\renewcommand{\arraystretch}{1.1}
\begin{tabular}{llcc}
\toprule
\textbf{Encoder} &
\textbf{Module} &
\textbf{Latency (ms)} &
\textbf{Energy ($\mu$J)} \\
\midrule
\multirow{3}{*}{\makecell{ResNet \\ ($d$=8, $s$=2)}}
& Bottleneck Encoder    & 123.42  & 33817.08 \\
& Quant+Huff & 0.06944 & 1.805 \\
& HDC   & 0.03242 & 0.729 \\
\midrule
\multirow{3}{*}{\makecell{SwinViT\\ ($d$=8, $s$=3)}}
& Attention Encoder    & 344.48  & 45421.06 \\
& Quant+Huff & 0.06827 & 1.775 \\
& HDC  & 0.03156 & 0.520 \\
\bottomrule
\end{tabular}
\end{table}

\begin{table}[t]
\caption{FPGA resource utilization and measured on-board dynamic power}
\label{tab:fpga_power}
\centering
\footnotesize
\setlength{\tabcolsep}{4pt}
\begin{tabular}{l r r r r c}
\toprule
\textbf{Hardware Module}
& \textbf{LUT}
& \textbf{FF}
& \textbf{BRAM}
& \textbf{DSP}
& \shortstack{\textbf{Dynamic}\\\textbf{Power (W)}} \\
\midrule
ResNet Encoder
& 20,942 & 20,943 & 177 & 66 & 0.277 \\

SwinViT Encoder$^{\dagger}$
& 10,529 & 13,576 & 257 & 31 & 0.128 \\

Quant + Huffman
& 5,031 & 4,960 & 2 & 0 & 0.025 \\

HDC
& 3,258 & 4,103 & 2 & 0 & 0.029 \\
\bottomrule
\multicolumn{6}{l}{
\scriptsize $^{\dagger}$Peak resource usage across separately implemented SwinViT blocks.
}
\vspace{-8mm}
\end{tabular}
\end{table}

\subsection{HDC Accuracy--Communication Trade-off}\label{sec:hdc_result}

We next examine the accuracy--communication trade-off of the HDC
classification path described in Section~\ref{sec:hdc_details} on the
VWW dataset. Since the energy, latency, and FPGA implementation costs
of HDC and Quant+Huff are analyzed in Section~VI-B, we focus here on
how the hypervector dimension $D$ and prototype retraining (RT) affect
classification accuracy and transmitted representation size. We
evaluate HDC using fixed encoder-latent configurations for both
architectures: $(d=8,s=3)$ for SwinViT and $(d=8,s=2)$ for ResNet.
For each encoder, we consider two hypervector dimensions and compare
the resulting accuracy and communication cost with the corresponding
Quant+Huff pipeline. The results are summarized in
Table~\ref{tab:hdc_results}.

Table~\ref{tab:hdc_results} reports the HDC classification results across two hypervector dimensions for each encoder, both without and with prototype retraining (RT). Without RT, the SwinViT-based encoder achieves $86.06\%$ accuracy at $D=128$, while the ResNet-based encoder achieves $89.78\%$ at $D=64$. Retraining uses misclassified samples to adjust the class prototypes, moving the correct prototype closer to the sample while pushing the incorrect prototype away, and is performed for up to 30 epochs with a batch size of 512. This improves the accuracy to $86.89\%$ for SwinViT ($D=128$) and $90.81\%$ for ResNet ($D=64$). 
%The communication reduction relative to the structured encoder latent is determined directly by $D$, yielding $2.25\times$ for SwinViT at $D=128$ and $2.00\times$ for ResNet at $D=64$ (Eq.~7). 
The communication reduction relative to the corresponding 4-bit encoder latent before Huffman coding is governed directly by $D$, yielding $2.25\times$ for SwinViT at $D=128$ and $2.00\times$ for ResNet at $D=64$
Interestingly, the accuracy impact of HDC is encoder dependent. For SwinViT, HDC provides additional communication reduction with a modest decrease in classification accuracy relative to the Quant+Huff path. In contrast, the ResNet-based HDC configuration with RT reaches $90.81\%$ accuracy, exceeding the $87.6\%$ accuracy of the corresponding $4$-bit Quant+Huff configuration reported in Table~VI, while simultaneously reducing the transmitted representation size. We attribute this difference to the stronger class separability of the compressed ResNet features after normalization and projection into the hypervector space, which allows the resulting class prototypes to remain more distinguishable under cosine-similarity-based inference. Thus, HDC should not be viewed solely as an accuracy--efficiency trade-off: depending on the encoder representation, it can provide both improved classification accuracy and lower communication cost.

Beyond these encoder-dependent accuracy characteristics, HDC offers several additional advantages for in-sensor classification. Prototype construction is gradient-free and relies on lightweight bundling operations, avoiding the iterative backpropagation required by a conventional classifier head. Likewise, encoding and associative-memory inference reduce to permute/bind/bundle/compare operations over binary or bipolar vectors rather than multi-bit MAC arithmetic. The distributed hypervector representation additionally provides intrinsic tolerance to bit-flip errors and hardware noise, while Sobol-based level generation improves hypervector orthogonality relative to conventional pseudo-random generation and contributes to the observed accuracy--dimension trade-off.

%From a hardware perspective, Table~\ref{tab:fpga_energy} shows that HDC achieves lower backend latency and energy than Quant+Huff for both encoder outputs. For ResNet, the backend energy decreases from $1.805~\mu$J with Quant+Huff to $1.575~\mu$J with HDC, corresponding to a $12.7\%$ reduction. For SwinViT, it decreases from $1.775~\mu$J to $1.208~\mu$J, corresponding to a $31.9\%$ reduction. Both post-encoder backends remain lightweight compared with the encoder accelerators. As shown in Table~\ref{tab:fpga_power}, Quant+Huff and HDC each consume less than $30$~mW and require only a small fraction of the overall hardware resources. Although HDC incurs a marginally higher dynamic power due to the additional hyperdimensional encoding operations, its lower execution latency results in reduced backend energy. Overall, these results show that HDC provides an efficient classification path with a small hardware footprint, while its accuracy and communication benefits depend on the characteristics of the encoder representation.

%In summary, quantization remains the preferred path when peak accuracy is the objective, whereas HDC offers a hardware-efficient, fault-tolerant alternative for classification workloads where single-pass training and logic-only inference are advantageous. This positions HDC not as a replacement for the default pipeline but as a complementary on-chip classification mode selectable according to the deployment's accuracy, energy, and robustness priorities.

\subsection{Ablation Study}
\label{sec:ablation}

To isolate the contribution of each component, we ablate several mechanisms that jointly produce OASIS's compression and accuracy on the 
VWW task with the SwinViT-based encoder at a fixed latent configuration 
($d=$8, $s=$3): (i) encoder-side 
dimensionality reduction, (ii) the entropy regularization loss, (iii) 
4-bit output quantization, (iv) Huffman coding, and (v) the 
HDC alternative in place of (iii)--(iv). Table~\ref{tab:ablation} 
reports test accuracy and the resulting bandwidth reduction as each 
component is added.

\begin{table}[t]
\centering
\caption{Ablation of OASIS components on VWW (SwinViT encoder). 
Bandwidth reduction is relative to the raw $224^2\times3$ input.}
\label{tab:ablation}
\footnotesize
\begin{tabular}{|l|c|c|}
\hline
\textbf{Configuration} & \textbf{Acc. (\%)} & \textbf{BW Red. ($\times$)} \\ \hline 
Encoder dim. reduction only            &  88.0 & 2,091 \\ \hline
\; + entropy loss ($\beta$)            &  87.9 & 2,091 \\ \hline
\; \quad + 4-bit quantization                & 87.6 & 4,181 \\ \hline
\; \quad \quad + Huffman coding (default path)       & 87.6 & 10,653 \\ \hline
\; \quad  + HDC Encoding (HDC alter. path) & 86.0 & 18,816 \\ \hline
 
\end{tabular}
\end{table}

\noindent\textbf{Effect of entropy regularization.} 
% [\,\textsc{fill: does $\beta$ change accuracy, and how much does it lower the 
% post-Huffman effective bit-width? compare $\beta{=}0$ vs $\beta{=}2$}\,]
Introducing the entropy regularization term has a negligible impact on the classification accuracy (88.0\%$\rightarrow$87.9\%) while preserving the same communication bandwidth. This indicates that the entropy objective mainly acts as a regularizer without introducing a measurable accuracy penalty in this setting.

\noindent\textbf{Effect of quantization bit-width.} 
% We additionally sweep the encoder output precision (Table~\ref{tab:bitwidth}) to identify the 
% accuracy--bandwidth knee. [\,\textsc{fill: at what bit-width does accuracy 
% start to drop materially? justify the 4-bit choice from this}\,]
We additionally sweep the encoder output precision (Table~\ref{tab:bitwidth}) to identify the accuracy--bandwidth knee. Reducing the precision from 8 bits to 4 bits incurs only a marginal accuracy drop (87.9\%$\rightarrow$87.6\%) while decreasing the post-Huffman effective bit-width from 3.77 to 1.57 bits. In contrast, further reducing the precision to 2 bits significantly degrades the accuracy to 70.3\%, despite only a modest additional reduction in effective bit-width (1.46 bits). These results identify 4-bit quantization as the best operating point, achieving nearly the maximum communication reduction while preserving almost all of the model accuracy.

\begin{table}[t]
\centering
\caption{Encoder output bit-width sweep on VWW (ResNet encoder).}
\label{tab:bitwidth}
\footnotesize
\begin{tabular}{|c|c|c|}
\hline
\textbf{Bit-width} & \textbf{Acc. (\%)} & \textbf{Eff. bits (post-Huffman)} \\ \hline
8 & 87.9 & 3.77 \\ \hline
4 & 87.6 & 1.57 \\ \hline
2 & 70.3 & 1.46 \\ \hline
\end{tabular}
\end{table}

% \noindent These results show that [\,\textsc{fill: 1-sentence takeaway — 
% which component drives compression, which drives accuracy retention}\,], 
% confirming that the gains are not attributable to any single mechanism but 
% to their combination.
% \noindent These results show that encoder-side dimensionality reduction is the primary contributor to accuracy retention, while quantization and Huffman
% coding are responsible for the majority of the communication savings, confirming that the final performance arises from the synergy of all four components rather than any single design choice.

\noindent\textbf{Effect of Huffman coding and the HDC alternative.} 
Huffman coding contributes the largest single jump in bandwidth reduction 
among all components (4,181$\times\rightarrow$10,653$\times$) with no 
additional accuracy cost, confirming its role as a near-lossless 
entropy-coding stage on top of quantization. Replacing quantization and 
Huffman coding with the HDC path instead yields a further bandwidth 
reduction (10,653$\times\rightarrow$18,816$\times$) at a modest accuracy 
cost (87.6\%$\rightarrow$86.0\%) on SwinViT-based encoder.

\noindent These results show that encoder-side dimensionality reduction 
is the primary contributor to accuracy retention, while quantization and 
Huffman coding are responsible for the majority of the communication 
savings within the default pipeline, and the HDC path offers a further 
bandwidth-accuracy trade-off point beyond it. This confirms that the 
final performance arises from the synergy of all components rather than 
any single design choice.

\begin{table}[t]
\centering
%\color{blue}
\caption{Inference accuracy and communication cost across HDC Dimension (VWW DATASET).}
\label{tab:hdc_results}
\setlength{\tabcolsep}{2.2pt}
\renewcommand{\arraystretch}{0.9}

\begin{tabular}{|l|r|r|r|r|}
\hline
\textbf{Encoder} & \textbf{Dim. (HDC)} & \textbf{Red. ($\times$)} & \textbf{Acc. w/o RT (\%)} & \textbf{Acc. w/ RT (\%)} \\
\hline
\multirow{2}{*}{SwinViT}
& 64  & 4.50 & 85.87 & 86.02 \\ \cline{2-5}
& 128 & 2.25 & 86.06 & 86.89 \\
\hline
\multirow{2}{*}{ResNet}
& 32 & 4.00 & 89.39 & 90.79 \\ \cline{2-5}
& 64 & 2.00 & 89.78 & 90.81 \\
\hline
\end{tabular}

\vspace{1mm}
\begin{minipage}{0.96\linewidth}
\scriptsize
RT denotes retraining of the HDC class prototypes using misclassified training samples after their initial construction.
\end{minipage}

\end{table}

%\begin{table}[t]
%\centering
%\caption{Comparison of classical and HDC-based Swin-Tiny models on the VWW task (5 epochs).}
%\label{tab:vww_hdc_vs_classical}
%\scriptsize
%\setlength{\tabcolsep}{1.2pt}
%\renewcommand{\arraystretch}{0.7}
%\begin{tabular}{|l|c|c|c|c|}
%\hline
%\textbf{Method} & \textbf{Acc. (\%)} & \textbf{Bandwidth (bits)} & \textbf{MIPI Energy ($\mu$J)} & \textbf{Energy/img (J)} \\
%\hline
%Swin+Linear & 93.42 & 12,288 & 0.0238 & 0.2273 \\
%\hline
%Swin+HDC, $D{=}1024$ & 86.40 & 1,024 & 0.0059 & 0.2249 \\
%\hline
%\end{tabular}
%\end{table}

\section{Conclusions \& Discussions}

We presented OASIS, a distributed in-sensor vision framework that
jointly optimizes task accuracy, latent entropy, and reconstruction
quality to generate compact representations before off-chip
transmission. OASIS supports a quantization-and-Huffman path for
classification and dense-prediction tasks, and a Sobol-based HDC path
for highly compressed classification. For the SwinViT-based VWW model, mapping the
$3\times3\times8$ latent representation to a 64-dimensional
binary hypervector provides an additional $1.77\times$ communication
reduction and an overall $18{,}816\times$ reduction relative to raw
8-bit image transmission. The quantization-and-Huffman path achieves up
to $11{,}985\times$ reduction while preserving structured features.
Across visual wake-word classification, hand tracking, and eye tracking,
OASIS reduces modeled total system energy by approximately
$2\times$--$4.5\times$ while maintaining competitive accuracy.

Because fabricating a heterogeneously integrated CIS and custom
near-sensor logic die is beyond the practical scope of this study, we
combine circuit simulation of the CIS front end, FPGA implementation and
board-level power measurements, Vivado post-implementation analysis, and
a 7-nm ASIC projection. Future work will explore adaptive compression,
emerging-memory integration, higher-resolution applications, and
multi-modal fusion across LiDAR, radar, and event-based sensors.
Overall, OASIS provides a practical path toward low-power in-sensor
vision for AR/VR, smart sensing, and autonomous systems.

%{\footnotesize  % Reduces space between bibliography items
%\bibliographystyle{unsrt}
\bibliographystyle{IEEEtran}
\bibliography{References}
%}

%\newpage

\section{Biography Section}

\begin{IEEEbiography}[{\includegraphics[width=1in,height=1.25in,clip,keepaspectratio]{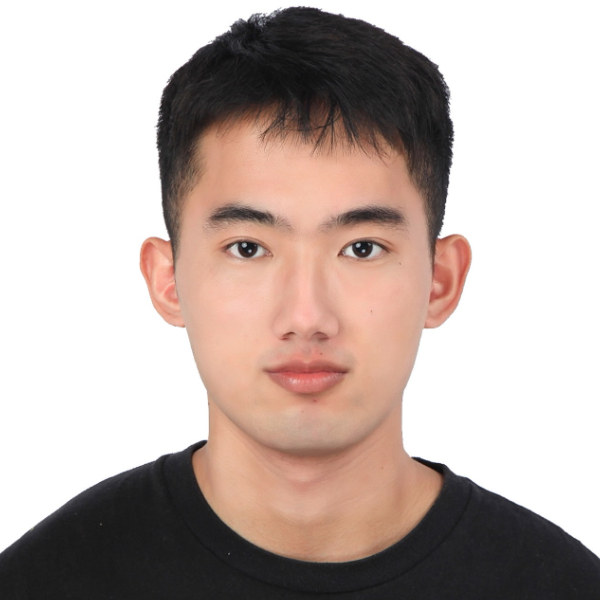}}]
{Chengwei Zhou} received his B.E. degree in Automation and M.Sc. degree in Electrical and Computer Engineering. He is currently a second-year Ph.D. student in Computer Engineering at Case Western Reserve University, Cleveland, USA, where he works as a graduate research assistant under the supervision of Dr. Gourav Datta. He was selected as a DAC 2026 Young Fellow and has published in top-tier conferences and journals, including ICML, DAC, ICCAD, ICASSP, BioCAS, GLSVLSI, ISVLSI, and IEEE TCAS-II. He actively serves as a reviewer for leading AI and EDA conferences, including ICLR, ICML (Golden Reviewer Award), NeurIPS, and ICCAD. His research interests include in-sensor computing, hardware–software co-design, and efficient machine learning models (including LLM/VLM) for edge intelligence. 
\end{IEEEbiography}

\begin{IEEEbiography}
[{\includegraphics[width=1.3in,height=1.25in,clip,keepaspectratio]{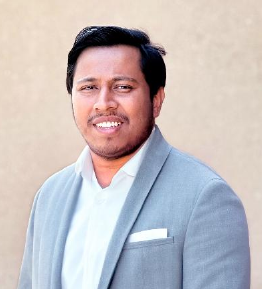}}]{Abu Kaisar Mohammad Masum} (S'25) received the B.Sc. degree in Computer Science and Engineering from Daffodil International University, Dhaka, Bangladesh, in 2020, and the M.S. degree in Computer Science from the University of Louisiana at Lafayette, Lafayette, LA, USA. He is currently a Ph.D. candidate in Computer Science at the University of Louisiana at Lafayette, specializing in quantum machine learning and emerging computing. He was selected as a DAC Young Fellow in 2025 and was a winner of the NASA Beyond the Algorithm Challenge. His research has been published in leading venues, including DAC, DATE, ISLPED, AAAI, TBME, TVLSI, Nature Portfolio, ESL, ISCAS, QCE, ICCD, GLSVLSI, and MWSCAS. His research interests include quantum computing, hyperdimensional computing, hardware-aware AI, and emerging computing systems. He currently serves as the Secretary of the IEEE Computer Society Lafayette Section and is a Student Member of the IEEE.
\end{IEEEbiography}

\begin{IEEEbiography}[{\includegraphics[width=1in,height=1.25in,clip,keepaspectratio]{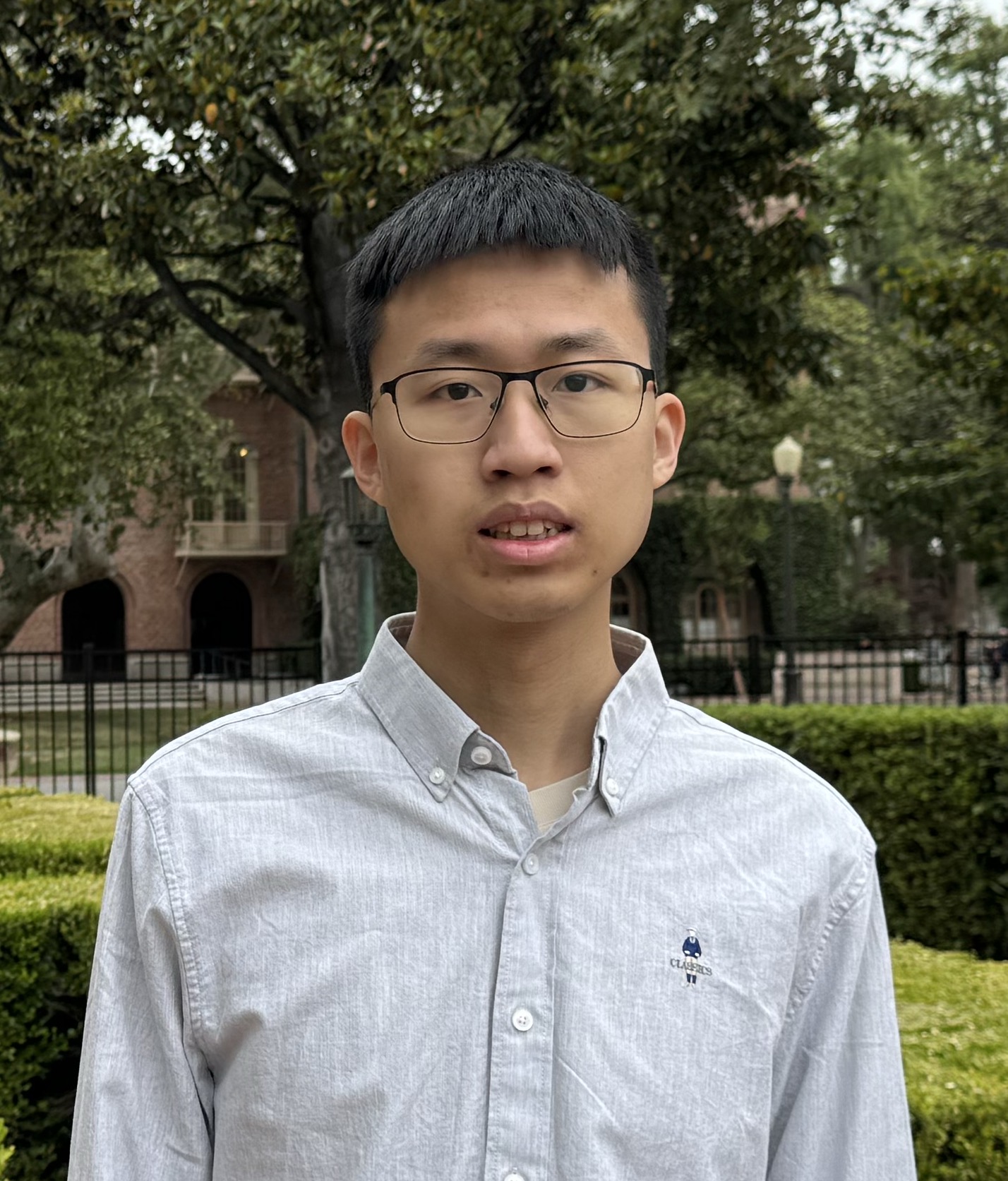}}]
{Xuming Chen} received the M.S. degree in Electrical and Computer Engineering from the University of Southern California, Los Angeles, CA, USA, in 2025. He is currently pursuing the Ph.D. degree in the Department of Electrical, Computer, and Systems Engineering at Case Western Reserve University, Cleveland, OH, USA, under the supervision of Prof.~Gourav Datta. He was selected as a DAC Young Fellow and has authored papers published at DAC and ICCAD in 2026. His research interests include algorithm--hardware co-design for efficient AI systems, hardware-aware machine learning, and emerging computing paradigms, with an emphasis on photonic computing, in-memory computing, and edge AI acceleration.
\end{IEEEbiography}

\begin{IEEEbiography}[{\includegraphics[width=1.3in,height=1.25in,clip,keepaspectratio]{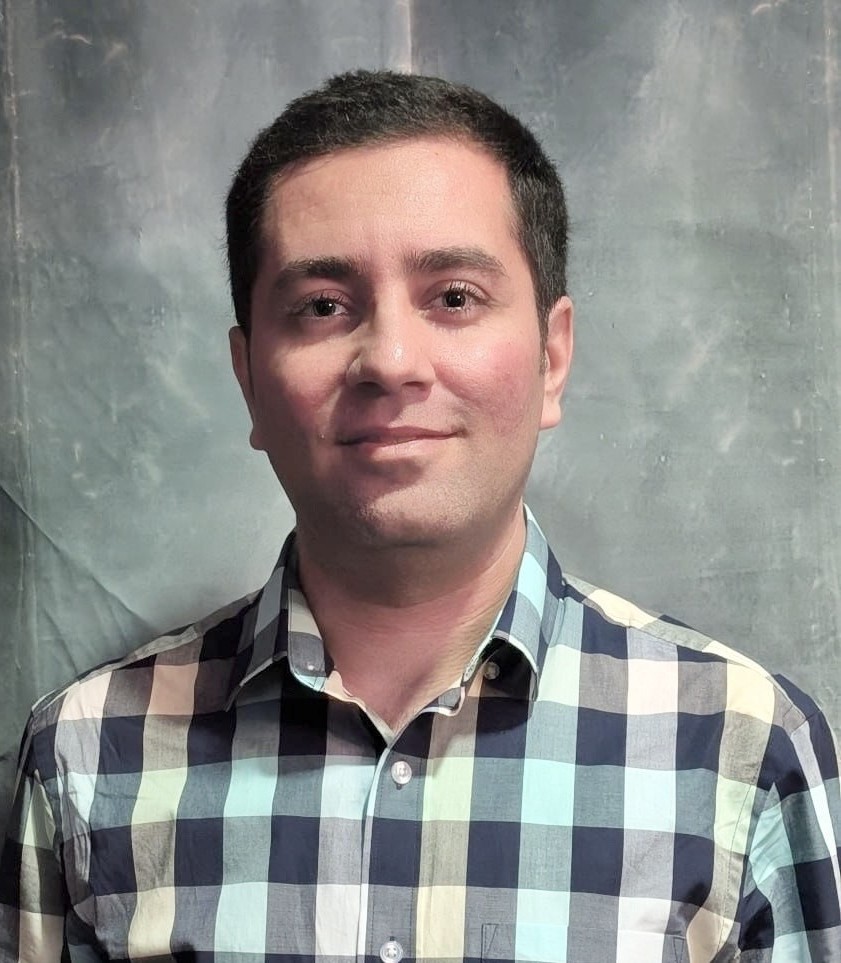}}]{Mehran Moghadam} (S’22) received the B.Sc. degree in Computer Engineering, and M.Sc. degree in \textit{Computer Systems Architecture} from the University of Isfahan, Iran, in 2010 and 2016. He graduated as one of the top-ranking students in both programs.
In 2022, he began his Ph.D. studies in Computer Engineering at the School of Computing and Informatics, University of Louisiana at Lafayette, Lafayette, LA, USA. In 2024, he transferred to the Department of Electrical, Computer, and Systems Engineering at Case Western Reserve University, Cleveland, OH, USA, to continue pursuing his Ph.D. in Computer Engineering.
He became a finalist in the ACM SIGBED Student Research Competition (SRC) at ESWEEK and ICCAD in 2024 and was selected as a Young Fellow in DAC 2024 and 2026. %He has authored/co-authored more than 30 peer-reviewed papers in top EDA venues.
Mehran received the Best Paper Award at ISLPED'26.
His research interests include emerging and unconventional computing paradigms, such as energy-efficient stochastic computing models, real-time and highly-accurate hyperdimensional computing systems, bit-stream processing, robust in-memory arithmetic computation, and low-power in/near-sensor computing designs for edge AI. 
\end{IEEEbiography}

\begin{IEEEbiography}[{\includegraphics[width=1in,height=1.25in,clip,keepaspectratio]{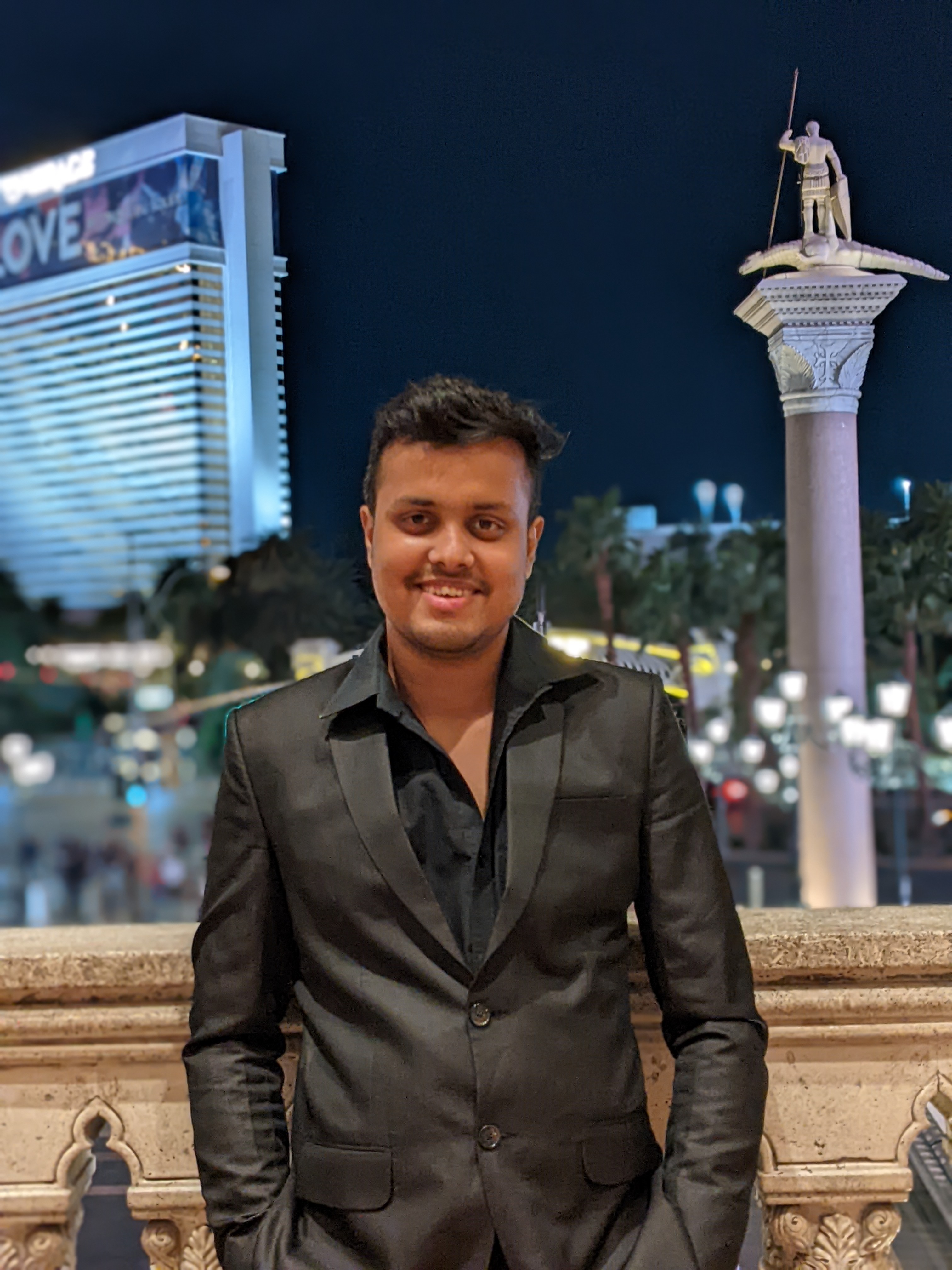}}]
{Arnab Sanyal} received a B.Tech. degree in Electrical Engineering and an M.Tech. degree in Instrumentation and Signal Processing from the Indian Institute of Technology Kharagpur, Kharagpur, WB, India, in 2017, and two M.S. degrees in Computer Science and Electrical Engineering from the University of Southern California, Los Angeles, CA, USA, in 2020. He is currently a Doctoral Student with the Department of Electrical and Computer Engineering at the University of Texas at Austin, Austin, TX, USA, and serves concurrently as a Senior Machine Learning Engineer at General Motors. His research has been published in leading venues, including IEEE ICASSP and ISVLSI. He has been adjudicated for an O-1 visa (Individuals with Extraordinary Ability or Achievement) by the United States government. His research interests include energy-efficient and robust machine learning algorithms for generative AI edge applications, approximate computations, and algorithm-architecture co-design for neural networks. 
\end{IEEEbiography}
%\vspace{-20mm}

\begin{IEEEbiography}[{\includegraphics[width=1in,height=1.25in,clip,keepaspectratio]{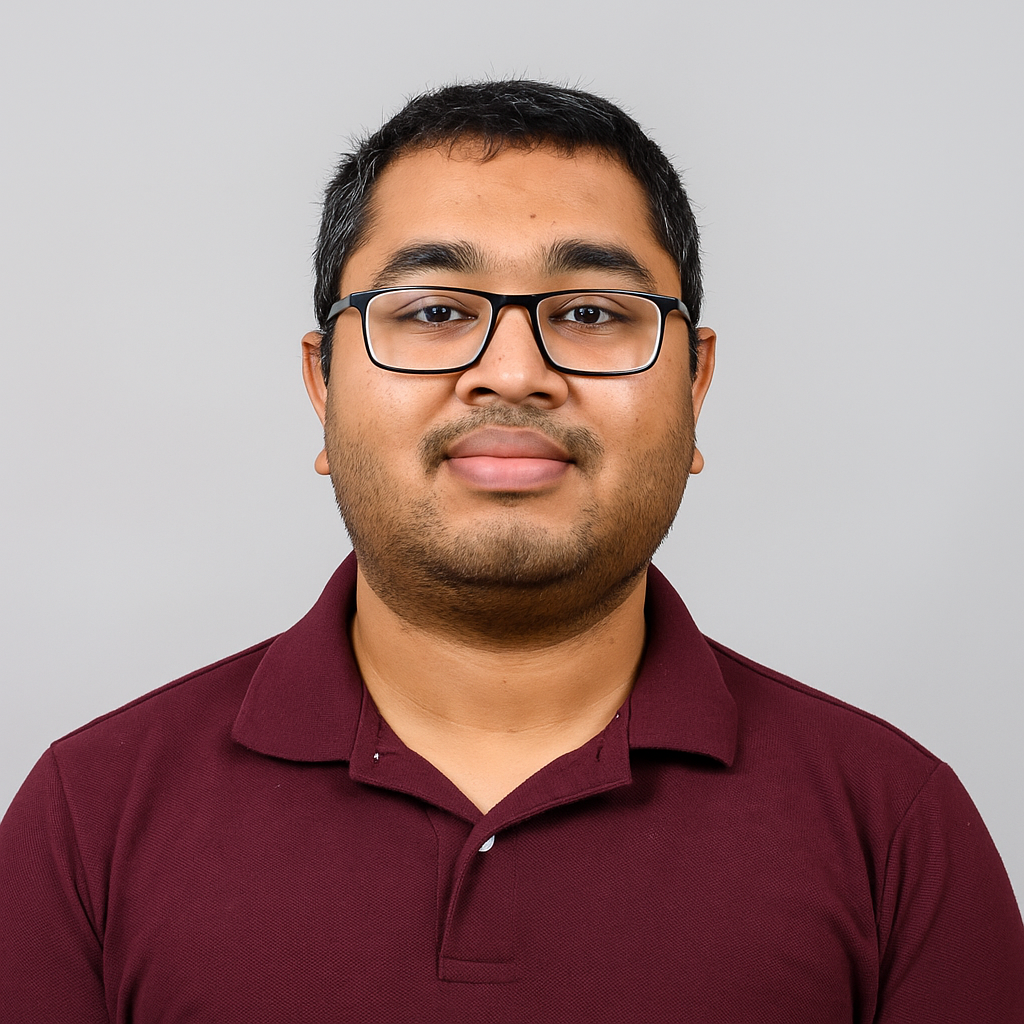}}]
{Md Abdullah-Al Kaiser} is an Analog IC Design Engineer at Apple. He earned his Ph.D. in Electrical and Computer Engineering from the University of Wisconsin–Madison in 2025. Prior to his Ph.D., he earned his M.S. in Electrical and Computer Engineering from the University of Southern California (2022). During his Ph.D., he received several prestigious honors, including the Harold Peterson Outstanding Dissertation Award, the DATE Travel Grant Award (2025), and the DAC Young Fellow Award (2024). He was also recognized with the Jenny Wang Excellence in Teaching Award (2024), the Charles Weber Teaching Award (2023), and the USC Annenberg Graduate Fellowship (2019–2023). Kaiser has authored more than 25 peer-reviewed publications in leading journals and conferences, including IEDM, the Design Automation Conference (DAC), npj Unconventional Computing, npj Nanophotonics, Frontiers, ICASSP, GLVLSI, and others. His research focuses on analog and mixed-signal integrated circuit design, in-memory and in-sensor computing, photonic computing, neuromorphic computing, and device-circuit-algorithm co-design to advance next-generation computing systems.
\end{IEEEbiography}

\begin{IEEEbiography}[{\includegraphics[width=1in,height=1.25in,clip,keepaspectratio]{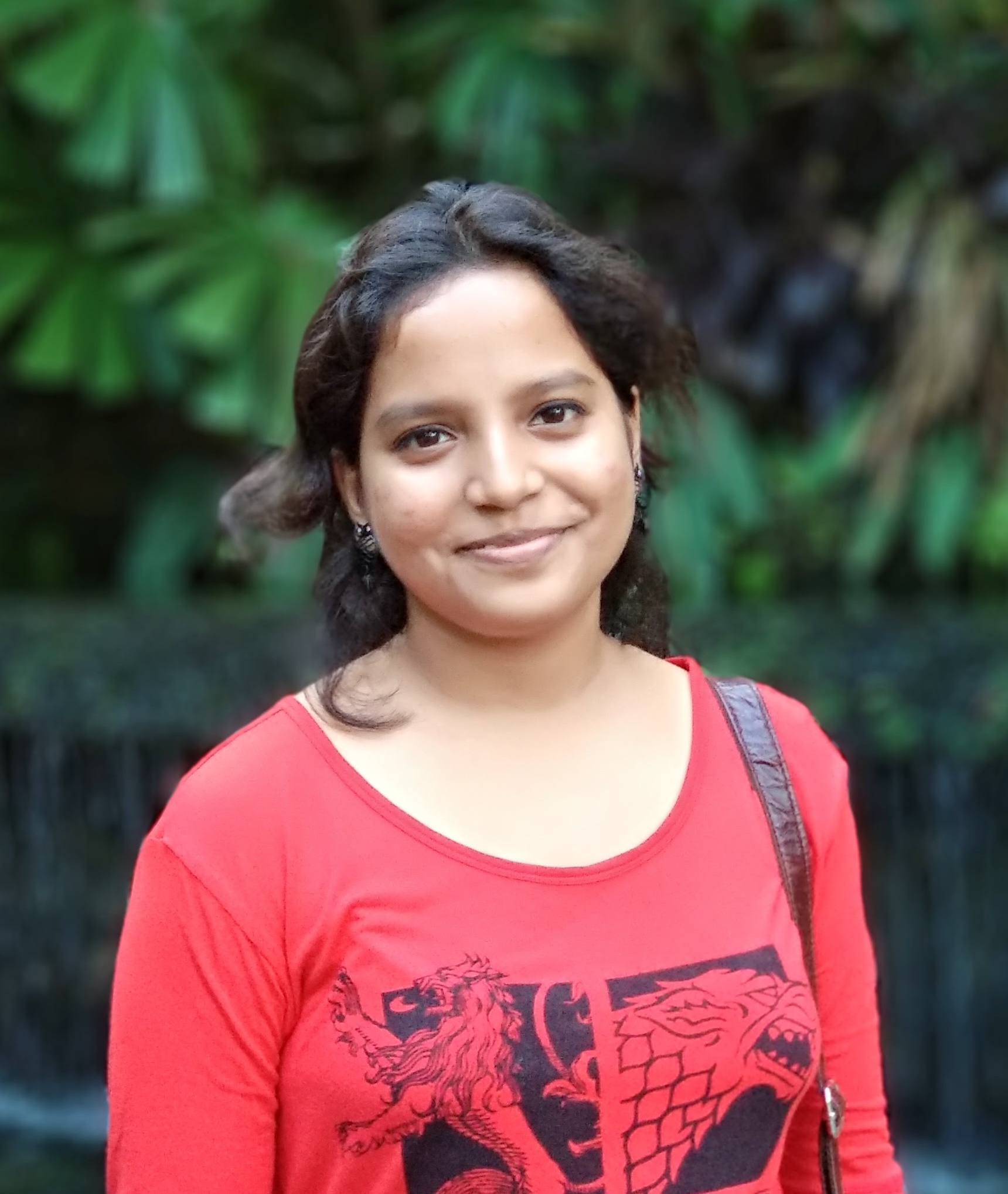}}]
{Sreetama Sarkar} is a fourth-year Ph.D. student in the Department of Electrical and Computer Engineering at the University of Southern California (USC), where her research focuses on energy efficiency and trustworthiness in multimodal foundation models. Before this, she completed a Master's degree from the Technical University of Munich and a Bachelor's degree from the National Institute of Technology Durgapur, India. During her Ph.D., she has completed research internships at Samsung Research America and Dolby Labs. She has authored 20+ peer-reviewed publications, including 8 first-author papers, in top venues such as EMNLP, CVPR, WACV, DAC, and ICCAD. Her work has been recognized through several honors at USC, including the WISE Graduate Merit Award (2026), the Annenberg Endowed Fellowship (2024), and the MHI Fellowship (2022). She received the 3rd place at the DAC Ph.D. Forum (2026) and has been invited to participate in the 13th Heidelberg Laureate Forum (2026).
\end{IEEEbiography}

\begin{IEEEbiography}[{\includegraphics[width=1.3in,height=1.25in,clip,keepaspectratio]{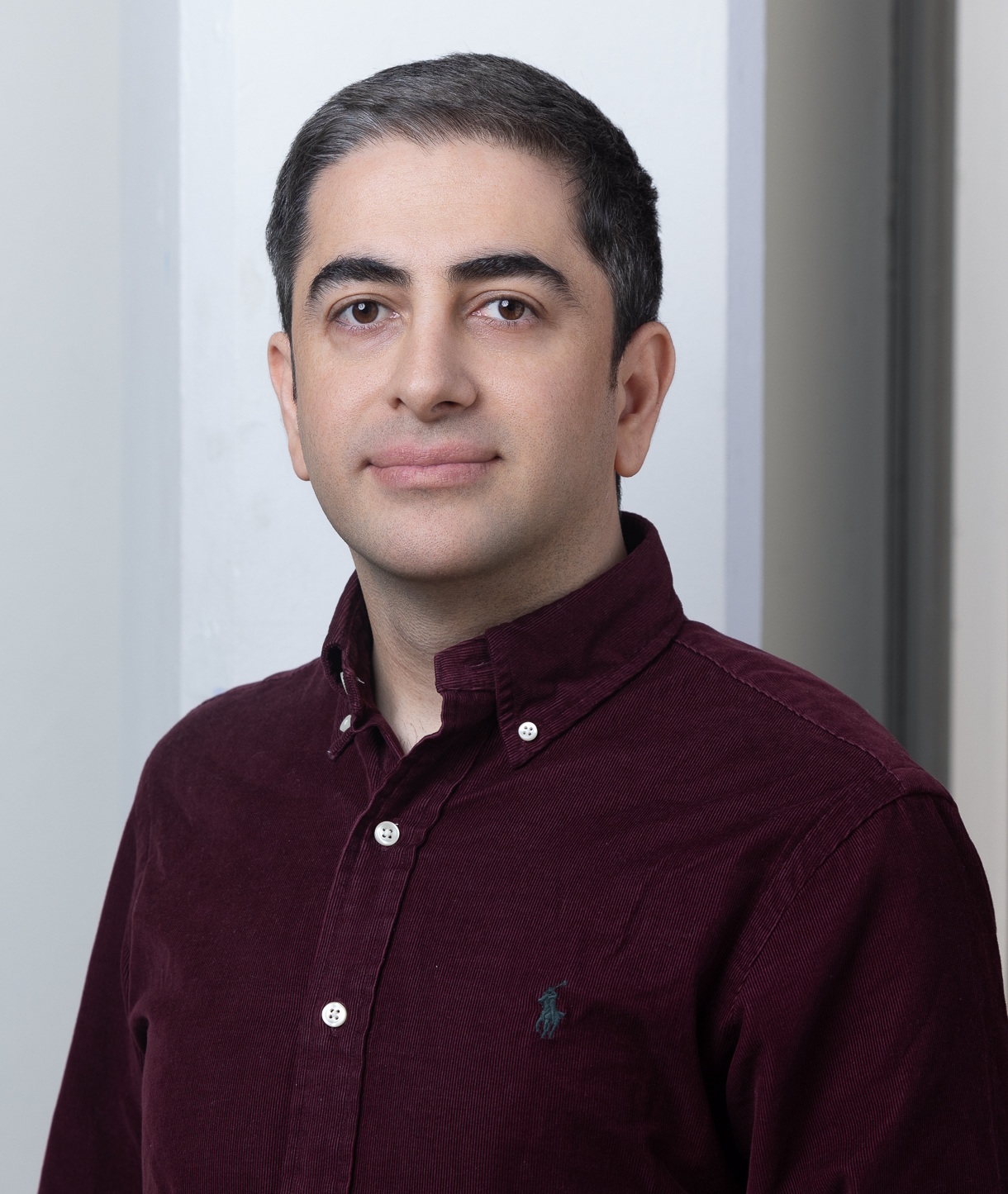}}]{M. Hassan Najafi} (S’15-M’18-SM'23) received the B.Sc. degree in Computer Engineering from the University of Isfahan, Iran, the M.Sc. degree in Computer Architecture from the University of Tehran, Iran, and the Ph.D. degree in Electrical Engineering from University of Minnesota, Twin Cities, USA, in 2011, 2014, and 2018, respectively. He was an Assistant Professor at the School of Computing and Informatics, University of Louisiana at Lafayette, Lafayette, LA, USA, from 2018 to 2024. He is currently an Associate Professor at the Electrical, Computer, and Systems Engineering Department at Case Western Reserve University, Cleveland, OH, USA. His research interests include stochastic and approximate computing, unary processing, in-memory computing, and hyperdimensional computing. He has authored/co-authored more than 120 peer-reviewed papers and has been granted 12 U.S. patents, with more pending. Dr. Najafi received the NSF CAREER Award in 2024, the Best Paper Award at GLSVLSI'23 and ICCD’17, the Best Poster Awards at GLSVLSI'24 and DCAS'26, the 2018 EDAA Outstanding Dissertation Award, and the Doctoral Dissertation Fellowship from the University of Minnesota. He has been an editor for the IEEE Journal on Emerging and Selected Topics in Circuits and Systems and a Technical Program Committee Member for many EDA conferences. Dr.~Najafi is a Senior Member of IEEE and a Senior Member of the U.S. National Academy of Inventors (NAI). \end{IEEEbiography}

\begin{IEEEbiography}[{\includegraphics[width=1.09in,height=1.2in]{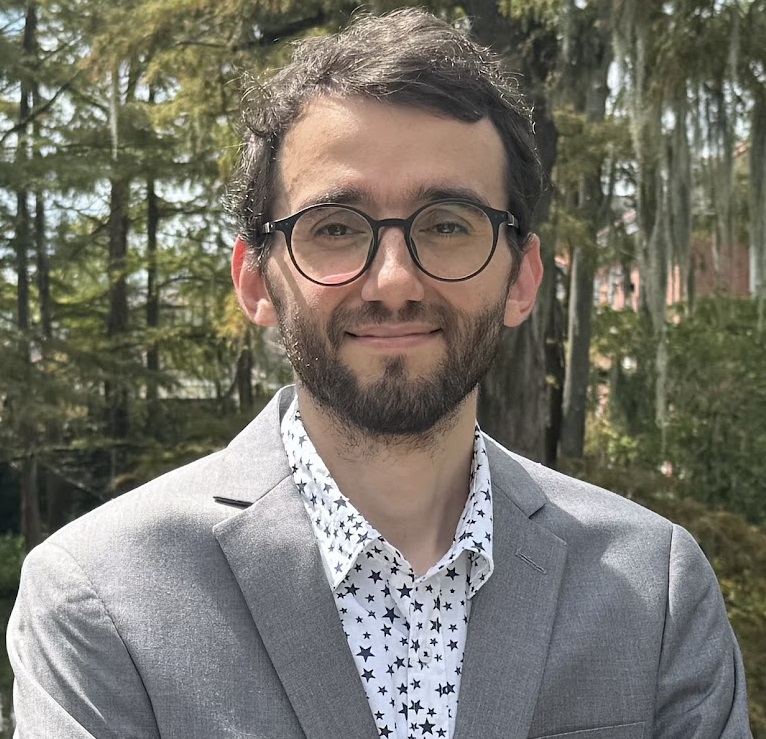}}]{Sercan Aygun} (S’09-M’22-SM'24) received a B.Sc. degree in Electrical \& Electronics Engineering and a double major in Computer Engineering from Eskisehir Osmangazi University, Turkey, in 2013. He completed his M.Sc. degree in Electronics Engineering from Istanbul Technical University in 2015 and a second M.Sc. degree in Computer Engineering from Anadolu University in 2016. Dr. Aygun received his Ph.D. in Electronics Engineering from Istanbul Technical University in 2022. %Dr. Aygun’s Ph.D. work has appeared in several Ph.D. Forums of top-tier conferences, such as DAC, DATE, ASP-DAC, and ESWEEK.
Dr. Aygun received the Best Scientific Research Award of the ACM SIGBED Student Research Competition (SRC) ESWEEK 2022, the Best Paper Award at GLSVLSI'23, and the Best Poster Awards at GLSVLSI'24 \& DCAS'26. Dr. Aygun's Ph.D. work was recognized with the Best Scientific Application Ph.D. Award by the Turkish Electronic Manufacturers Association and was also ranked first nationwide in the Science and Engineering Ph.D. Thesis Awards by the Turkish Academy of Sciences. He is currently an Assistant Professor at the School of Computing and Informatics, University of Louisiana at Lafayette, Lafayette, LA, USA. He works on tiny machine learning and emerging computing, including stochastic and hyperdimensional computing. \end{IEEEbiography}
%\vspace{-20mm}

\begin{IEEEbiography}[{\includegraphics[width=1in,height=1.25in,clip,keepaspectratio]{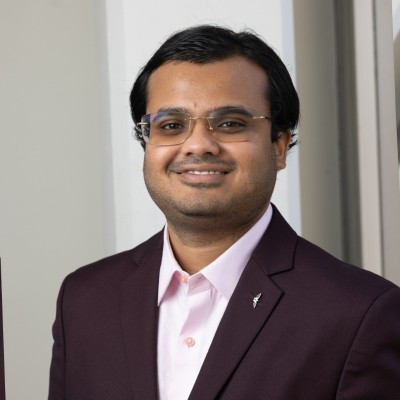}}]
{Dr. Gourav Datta} is an Assistant Professor with the ECSE Department of Case Western Reserve University (CWRU). Before that, he was an Applied Scientist at Amazon AGI, where he contributed to enhancing the video understanding capabilities of Amazon Nova, a next-generation foundation model. He obtained his PhD degree from University of Southern California in 2023. His research interests include energy and latency efficient algorithm-hardware co-design for machine learning at the edge. During his PhD tenure, he received the USC Viterbi School of Engineering’s highest achievement award, The 2024 William Ballhaus Best PhD Dissertation Award, the IEEE Graduate Fellowship on Applied Superconductivity 2022, the best Research Assistant (RA) award from USC ECE, and the Annenberg fellowship. He has published more than 60 peer-reviewed papers in top-tier venues including Nature Scientific Reports, Frontiers in Neuroscience, ICLR, ECCV, TCAS-I, DATE, WACV, among others, and received a best paper award nomination at VLSI-SoC 2022. Lastly, his research on in-sensor computing has been highlighted multiple times by Edge Impulse, a leading semiconductor IP company for TinyML, and USC Viterbi School of Engineering. 
\end{IEEEbiography}

\vfill

\end{document}